\documentclass[conference]{IEEEtran}
\IEEEoverridecommandlockouts
\usepackage{graphicx} 
\usepackage{epstopdf}
\usepackage{float}
\usepackage{amsmath}
\usepackage{amsthm} 

\newtheorem{myDef}{Definition}

\usepackage{stfloats}
\usepackage{cite}
\usepackage{subfigure}
\usepackage{amsmath,amssymb,amsfonts}
\usepackage{algorithmic}
\usepackage{graphicx}
\usepackage{textcomp}
\usepackage{verbatim}
\usepackage{multirow}
\usepackage{hyperref}
\usepackage{marvosym}
\usepackage{booktabs}
\hypersetup{hypertex=true,
	colorlinks=true,
	linkcolor=blue,
	anchorcolor=blue,
	citecolor=blue}
\usepackage{xcolor}
\usepackage{algorithmic}
\usepackage{bm}
\usepackage[ruled,vlined]{algorithm2e}
\def\BibTeX{{\rm B\kern-.05em{\sc i\kern-.025em b}\kern-.08em
		T\kern-.1667em\lower.7ex\hbox{E}\kern-.125emX}}
\begin{document}
\title{GA-DRL: Graph Neural Network-Augmented Deep Reinforcement Learning for DAG Task Scheduling over Dynamic Vehicular Clouds \\
}

\author{\IEEEauthorblockN{Zhang Liu, \emph{Student Member, IEEE}, Lianfen Huang, \emph{Member, IEEE}, Zhibin Gao, \emph{Member, IEEE},\\ Manman Luo, \emph{Student Member, IEEE}, Seyyedali Hosseinalipour, \emph{Member, IEEE}, Huaiyu Dai, \emph{Fellow, IEEE}}
	
	\thanks{
		
		\emph{Z. Liu (zhangliu@stu.xmu.edu.cn) and L. Huang (lfhuang@xmu.edu.cn) are with the Department of Informatics and Communication Engineering, Xiamen University, Fujian, China. S. Hosseinalipour (alipour@buffalo.edu) is with the Department of Electrical Engineering, University at Buffalo-SUNY, Buffalo, NY 14260. M. Luo (luomanman@stu.xmu.edu.cn) is with the Department of Electronic Engineering, Xiamen University, Xiamen, China. Z. Gao (gaozhibin@jmu.edu.cn) is with the Navigation Institute, Jimei University, Xiamen, Fujian, China. H. Dai (hdai@ncsu.edu) is with the Department of Electrical and Computer Engineering, North Carolina State University, Raleigh, NC 26795 USA. (Corresponding author: Lianfen Huang).} 
	}
}

\maketitle
\begin{abstract}
Vehicular clouds (VCs) are modern platforms for processing of computation-intensive tasks over vehicles. Such tasks are often represented as directed acyclic graphs (DAGs) consisting of interdependent vertices/subtasks and directed edges. In this paper, we propose a \underline{g}raph neural network-\underline{a}ugmented \underline{d}eep \underline{r}einforcement \underline{l}earning scheme (GA-DRL) for scheduling DAG tasks over dynamic VCs. In doing so, we first model the VC-assisted DAG task scheduling as a Markov decision process. We then adopt a multi-head \underline{g}raph \underline{at}tention network (GAT) to extract the features of DAG subtasks. Our developed GAT enables a two-way aggregation of the topological information in a DAG task by simultaneously considering predecessors and successors of each subtask. We further introduce non-uniform DAG neighborhood sampling through codifying the scheduling priority of different subtasks, which makes our developed GAT generalizable to completely unseen DAG task topologies. Finally, we augment GAT into a double deep Q-network learning module to conduct subtask-to-vehicle assignment according to the extracted features of subtasks, while considering the dynamics and heterogeneity of the vehicles in VCs. Through simulating various DAG tasks under real-world movement traces of vehicles, we demonstrate that GA-DRL outperforms existing benchmarks in terms of DAG task completion time.
\end{abstract}

\begin{IEEEkeywords}
Vehicular cloud, directed acyclic graph, deep reinforcement learning, graph neural network.
\end{IEEEkeywords}

\section{Introduction}
\subsection{Background and Challenges}	
\begin{figure}[htbp]
	\centering
	\includegraphics[width=3.5in]{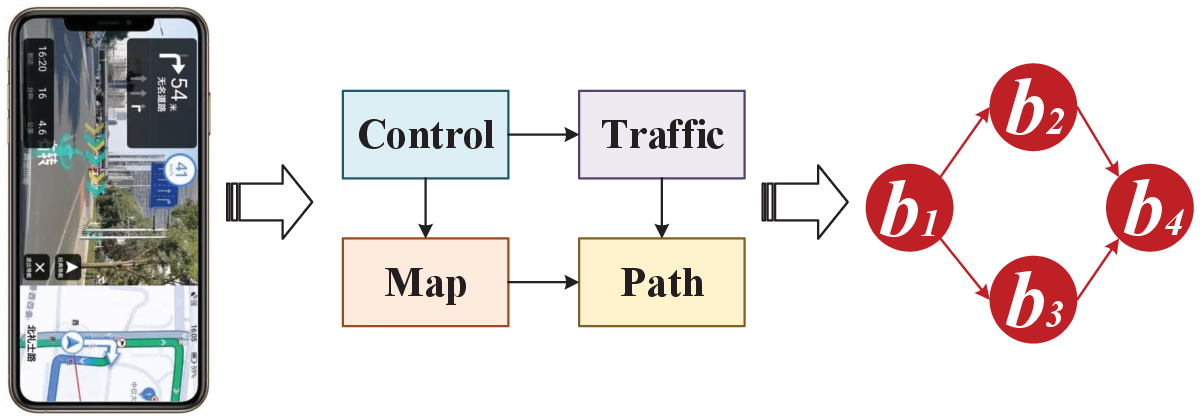}
	\caption{A schematic of a DAG task describing a navigation application\cite{b27}.}
	\label{fig_1}
\end{figure}

Vehicular networks are one of the main components of the Internet-of-Things (IoT) ecosystem. They have been envisioned to provide a reliable platform for execution of a myriad of applications/tasks, such as autonomous driving and mobile E-health\cite{b1},\cite{b2}. Many of these tasks possess complex computation topologies, which are often represented as directed acyclic graphs (DAGs) \cite{b28}. Fig. \ref{fig_1} illustrates a real-world DAG task model corresponding to a navigation application executed on a vehicle\cite{b27}, where vertices denote subtasks of the task and directed edges describe the dependencies between the execution of subtasks. In particular, each subtask represents a \emph{processing component} of navigation, while directed edges dictate the \emph{sequence of executions} of subtasks. The sequential execution of subtasks in a DAG model stems from the fact that processing of a subtask may  depend on the output data of others (e.g., in Fig. \ref{fig_1}, processing of subtask $b_2$ relies on the output data of subtask $b_1$ and processing of $b_4$ relies on the output data of both $b_2$ and $b_3$). 

In vehicular networks, DAG tasks are frequently encountered. Nevertheless, one of the main obstacles in the execution of DAG tasks is that a task owner (i.e., a vehicle with a DAG task) in a vehicular network often fails to fulfill the task's execution requirements due to its limited on-board resources. To circumvent this, offloading the computation of DAG tasks from task owners to edge servers through the mobile edge computing (MEC) platform has been proposed\cite{b12,b13,b14}. However, such task offloading strategies often rely on vehicle-to-infrastructure (V2I) communications, which can suffer from a high latency (e.g., due to a high data traffic congestion on the fronthaul/backhaul links) and limited coverage (e.g., in suburban areas)\cite{b9}. In response to these limitations, vehicular clouds (VCs) have emerged as novel computing platforms that integrate heterogeneous and distributed computation resources of moving vehicles via opportunistic vehicle-to-vehicle (V2V) communications to build flexible and scalable computing topologies for real-time task processing\cite{b15,b16,b17}. Specifically, in a VC, DAG subtasks are dispersed across vehicles and the data needed for the execution of subtasks is transmitted via V2V links.
 
Although DAG task processing over VCs is promising, efficient scheduling of DAG subtasks across vehicles is a highly non-trivial problem, which resembles mixed integer programming (MIP) due to the existence of continuous and binary variables in the formulation (detailed in Section III-E). MIP are NP-hard problems\cite{b43}, for which, dynamic programming \cite{b4}, \cite{b7} and list scheduling algorithms\cite{b5},\cite{b12} have been widely used to obtain solutions. These algorithms, however, often suffer from a prohibitively high computation complexity, which renders them impractical for large-scale VC networks. Also, these algorithms require a prior knowledge about the system dynamics (e.g., time-varying V2V channel qualities), which is cumbersome to acquire in practical systems. To overcome these challenges, researchers have recently started exploring the learning-based methods, a popular example of which is deep reinforcement learning (DRL)\cite{b20,b21,b22,b23,b24,b25,b26}. Roughly speaking, DRL learns from interacting with an environment so as to generate real-time near-optimal mappings from the state space (detailed in Section IV-C) to the action space (detailed in Section IV-D) without requiring any prior knowledges about the environment.

Although DRL has shown a tremendous success in task scheduling/offloading\cite{b20,b21,b22,b23,b24,b25,b26}, it cannot be readily adopted for scheduling of DAG tasks over dynamic VCs. This is due to the fact that DAG tasks' data (i.e., tasks' topologies) resides in a non-Euclidean space (i.e., graph). As a result, conventional DRL with handcrafted states designed to work with data in Euclidean spaces fails to automatically learn the topological information of DAG tasks, and thus can be hardly applied to unseen DAG task topologies upon deployment in real-world systems. To overcome this challenge, we propose to augment DRL with an emerging learning architecture called graph neural network (GNN). GNNs are capable of adaptively extracting discriminative features for each node of a graph based on the topological information aggregated from its neighboring nodes\cite{b41}. As a sub-category of GNNs, graph attention networks (GATs) have recently gained tremendous attentions, which extend the spatial convolution in convolutional neural networks (CNNs) to graph structures \cite{b42} and thus enjoy \emph{inductive learning}, making their learned models generalizable to unseen graph topologies.

\subsection{Overview and Summary of Contributions}
Inspired by the unique advantages of DRL and GNNs, we propose GA-DRL, a GNN-augmented DRL scheme to conduct DAG subtask-to-vehicle allocation aiming at minimizing the DAG task completion time. In doing so, we first model the VC-assisted DAG task scheduling as a Markov decision process (MDP). We then tailor a GAT to extract a set of features for each subtask of a DAG task. Finally, we integrate our developed GAT into the learning architecture of a double deep Q-network (DDQN) to generate subtask-to-vehicle allocation decisions, while taking into account the dynamics and heterogeneity of vehicles in a VC. Major contributions of this paper can be summarized as follows:
\begin{itemize}
	\item We develop a multi-head GAT capable of extracting features for DAG subtasks. Particularly, our GAT conducts a two-way topological information aggregation by simultaneously considering predecessors and successors of each subtask. Further, we incorporate a non-uniform neighborhood sampling methodology into our GAT by codifying the scheduling priority of subtasks, making our GAT generalizable to unseen DAG task topologies upon being deployed over the real-world systems.
	\item We propose a DDQN to conduct subtask-to-vehicle allocation decisions according to the extracted features of subtasks by our GAT, while taking into account the dynamics and heterogeneity of vehicles in a VC. Further, we incorporate an action mask module into the DDQN to avoid infeasible subtask-to-vehicle allocations, ensuring successful execution of subtasks. 
	\item We evaluate the performance of GA-DRL on a real-world road network obtained from OpenStreetMap\cite{b10} where SUMO\cite{b11}, one of the most popular softwares for generating traffic flow, is used to form a VC. Through simulating DAG tasks with various topologies, we reveal that GA-DRL can outperform existing benchmarks in terms of task completion time.
\end{itemize}

The rest of this paper is organized as follows: Section II contains the related work. In Section III, we present the system model and formulate the VC-assisted DAG task scheduling as a MIP problem. We develop GA-DRL in Section IV. In Section V, we present simulation results before concluding the paper in Section VI.
\section{Related Work}
Existing works on DAG task scheduling over cloud-assisted networks can be roughly divided into two categories with respect to the type of their scheduling mechanisms: \emph{i}) heuristic-based algorithms \cite{b3},\cite{b4},\cite{b6},\cite{b12,b13,b14,b15},\cite{b17}; \emph{ii}) learning-based methods\cite{b16},\cite{b18},\cite{b20,b21,b22,b23,b24,b25,b26}. Below, we summarize the contributions of these works, and highlight the differences between our methodology in this paper and prior works.
\subsection{Heuristic DAG Task Scheduling}
\subsubsection{Static computing environment}
Heuristic methods for DAG task scheduling have been extensively studied for static MEC networks with fully connected servers\cite{b3},\cite{b4},\cite{b6},\cite{b12,b13}. H. Topcuoglu \emph{et al.} in \cite{b12} proposed HEFT algorithm, where each subtask is assigned to the processor with the least execution time. In \cite{b3}, L. F. Bittencourt \emph{et al.} proposed forward looking attributions to improve the performance of HEFT. In \cite{b6}, H. Kanemitsu \emph{et al.} proposed a clustering-based DAG task scheduling algorithm via prioritizing assigning the subtasks located on the critical path to the same processor. G. C. Sih \emph{et al.} in \cite{b4} adopted a compile-time-aware scheduling algorithm to dynamically allocate DAG subtasks over the existing processing units in the system. Recently, in \cite{b13}, Y. Sahni \emph{et al.} introduced JDOFH to simultaneously consider dependencies among DAG subtasks and start time of network flows to transmit the data of subtasks over the network.
\subsubsection{Dynamic computing environment}
Few recent works have studied DAG task scheduling over dynamic networks\cite{b14},\cite{b15},\cite{b17}. Q. Shen \emph{et al.} in \cite{b14} proposed DTOSC to conduct DAG task offloading and service caching in vehicular edge computing. F. Sun \emph{et al.} in \cite{b15} addressed DAG task scheduling over VC via a modified genetic algorithm focusing on vehicles' dwell
times. In \cite{b17}, Y. Liu \emph{et al.} developed MAMTS to prioritize allocation of different DAG tasks according to their computation topologies in vehicular edge computing. 

The methodologies developed in the aforementioned works are heuristic, applying of which requires considerable number of iterations to reach locally optimal solutions. As a result they often suffer from prohibitively high computation complexities, which renders them impractical for real-time DAG task allocation. Also, these heuristic algorithms often presume a prior knowledge about the system dynamics (e.g., known time-varying V2V channel qualities), obtaining of which is extremely challenging in dynamic VCs, where the network topology may exhibit a significant temporal variation. 
\vspace{-0.5em}
\subsection{Learning-based DAG Task Scheduling}
\subsubsection{Static computing environment}
DRL schemes have become one of the most popular learning-based techniques in the literature of task scheduling, especially for static MEC networks\cite{b20,b21,b22,b23,b24}. In \cite{b20}, J. Yan \emph{et al.} proposed an actor-critic DRL to learn the optimal DAG subtask assignment to access points. M. S. Mekala \emph{et al.} in \cite{b21} developed a DRL-based DAG task offloading approach to reduce the utilization cost of edge servers. In\cite{b22}, J. Wang \emph{et al.} proposed a DAG task offloading methodology based on meta reinforcement learning. M. Goudarzi \emph{et al.} in\cite{b23} introduced weighted actor-learner architectures for DAG task allocation over resource-constrained IoT devices. In\cite{b24}, Z. Hu \emph{et al.} presented a DRL-based Monte-Carlo tree search method to minimize DAG tasks' completion times through a clustered scheduler. 

\subsubsection{Dynamic computing environment}
Considering dynamic computing environments\cite{b16},\cite{b18}, \cite{b25,b26},  H. Liu \emph{et al.} in \cite{b16} utilized a policy-based DRL for minimizing DAG tasks' completion times in multi-vehicle scenarios. In \cite{b18}, J. Shi \emph{et al.} proposed a DRL-based DAG task offloading scheme for vehicular fog computing considering the vehicles' mobility and availability. X. Wei \emph{et al.} in \cite{b25} developed a DRL-based algorithm to jointly optimize the unmanned aerial vehicle trajectory planning, and DAG task scheduling. In \cite{b26}, L. Geng \emph{et al.} proposed a multi-agent actor-critic DRL to schedule DAG tasks in a vehicular edge computing network. 

The DRL algorithms developed in the above works are based on handcrafted features, making them unable to fully capture the existing topological information in DAG tasks. This is because the state space of the DRL architectures studied in the above works merely contains basic, human-selected information regarding subtasks (e.g., their computation workloads, transmission data sizes, and number of predecessors/successors). As a result, the DRL methods explored in the above works are solely capable of making allocation decisions for DAG tasks with computation topologies that they have seen during their training period. In this work, we take the first steps towards addressing this limitation.

\subsection{Footprints of GNNs in Mobile Edge Computing}
Recently, the success of GNNs in solving a variety of complex problems in wireless communications has been revealed \cite{b34,b35,b36,b37}, while studying their application in the context of DAG task scheduling is still in early stages. In \cite{b34}, Z. He \emph{et al.} investigated the spectrum allocation in vehicle-to-everything networks based on the integration of GNNs and deep Q-learning. Y. Li \emph{et al.} in \cite{b35} proposed a meta-reinforcement learning method for DAG task offloading in MEC platform, where the interdependencies between subtasks was extracted by GNNs. In \cite{b36}, H. Lee \emph{et al.} developed a graph convolution network (GCN) and DRL to effectively learn a priority-based scheduling policy for DAG tasks. J. Chen \emph{et al.} in \cite{b37} proposed an algorithm called ACED for DAG task offloading, where a GCN is leveraged to capture the topological information of DAG subtasks.

The aforementioned works either ignore the topology of computation-intensive tasks (e.g., interdependencies among subtasks)\cite{b34} or focus on static MEC environments, overlooking the dynamics and instability of resource provisioning\cite{b35,b36,b37}, which are significant features of VCs. Moreover, the GCN architecture developed in \cite{b36,b37} relies on \emph{transductive learning}, which requires knowing the graph structure of DAG tasks upfront. As a result, their learned solutions for DAG task scheduling are not applicable to unseen DAG task topologies, which makes them suffer from a prohibitively high training overhead for each newly arrived DAG task to the system. In this work, we are particularly interested to address the shortcomings mentioned above.

\section{System Model and Problem Formulation}
In this section, we first give an overview of the system of our interest, DAG task model, vehicle mobility model, and computation offloading model. We then obtain an optimization formulation for VC-assisted DAG task scheduling. Table \ref{table1} summarizes the major notations used in this section. 
\vspace{-0.5em}
\subsection{System Overview}
We consider a time-slotted VC-assisted DAG task scheduling scenario, which is coordinated by a road side unit (RSU) with coverage diameter of $D$. We presume that the area comprises $|\mathcal{V}|$ vehicles collected by the set $\mathcal{V}=\{v_m \mid 1 \leq m \leq |\mathcal{V}|\}$. In order to fulfill its DAG task completion demands, a task owner engages in offloading its DAG task with $|\mathcal{B}|$ subtasks collected by the set $\mathcal{B}=\{b_i \mid 1 \leq i \leq |\mathcal{B}|\}$ to other vehicles\footnote{This paper investigates the DAG task scheduling problem for a single task owner with a single DAG task in one VC for analytical simplicity. Cooperations and resource sharing among VCs and competitions between multiple task owners to acquire computation resources are left as future work.}. 
\begin{figure}[!t]
	\centering
	\includegraphics[width=3.5in]{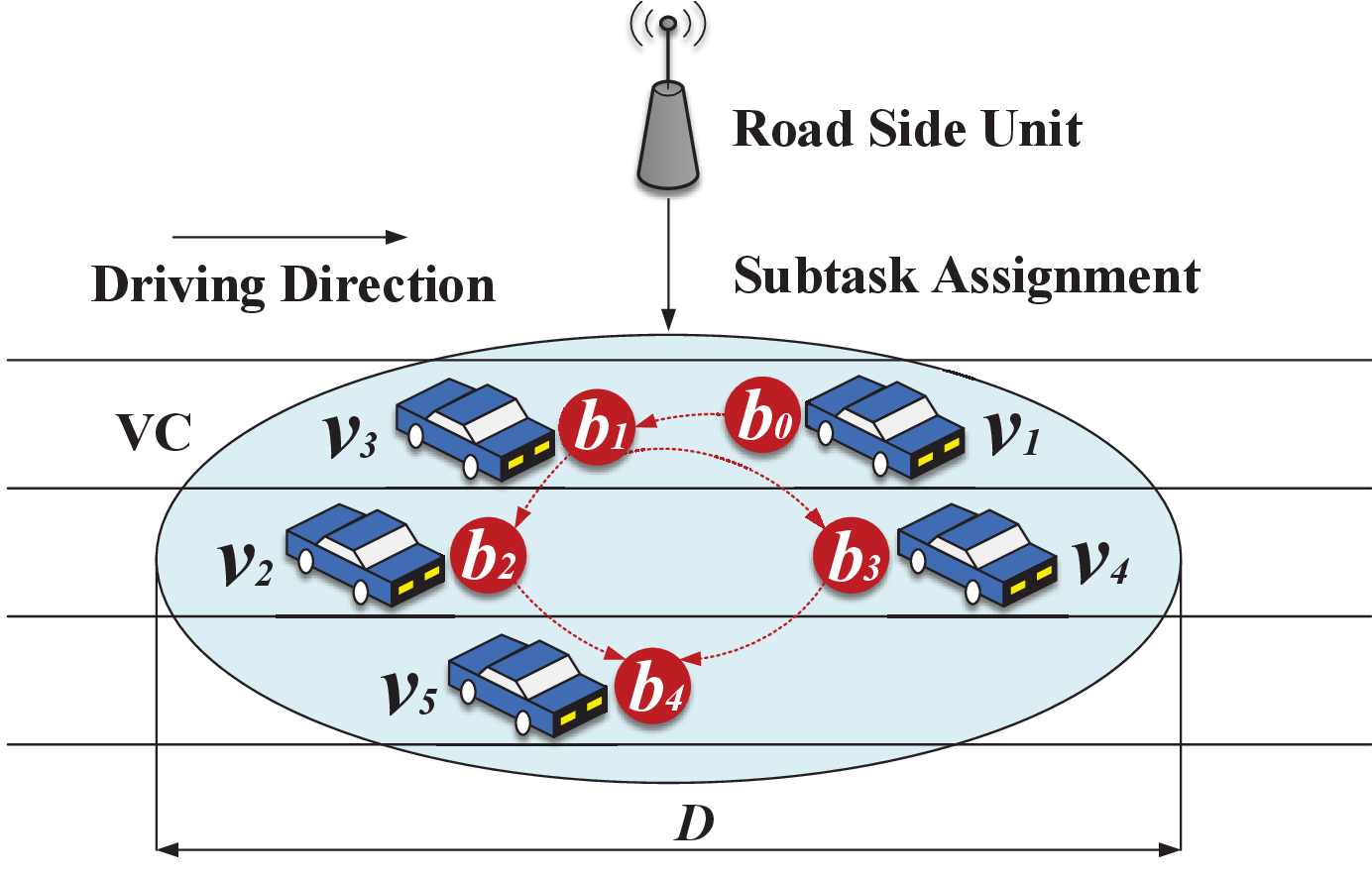}
	\caption{VC-assisted cooperative DAG task scheduling scenario.}
	\label{fig_2}
	\vspace{-0.5cm}
\end{figure}

Fig. \ref{fig_2} shows a schematic of our VC of interest for the DAG task topology depicted in Fig. \ref{fig_1}, subtask $b_0$ is a virtual subtask executed on the task owner (detailed in Section III-B). After receiving the offloading request from the task owner (i.e., $v_1$), the RSU acts as a centralized coordinator\cite{b16}, which processes a set of collected data (e.g., locations and resources of vehicles) to assign DAG subtasks to vehicles. Specifically, in Fig. \ref{fig_2}, virtual subtask $b_0$ is assumed to be executed on the task owner locally, while subtask $b_{1}$ is allocated to vehicle $v_3$. After executing subtask $b_{1}$, vehicle $v_{3}$ is scheduled to transmit the output data of subtask $b_{1}$ to vehicles $v_{2}$ and $v_4$ for processing subtasks $b_{2}$ and $b_{3}$. Due to the interdependencies among DAG subtasks, the execution of subtask $b_4$ relies on the output data of both subtasks $b_2$ and $b_3$. Hence, vehicles $v_{2}$ and $v_{4}$ will both be scheduled to transmit their output data to vehicle $v_5$. Finally, vehicle $v_5$ will send a feedback (i.e., the final result of DAG task processing) to the RSU. The main assumptions made in this paper are summarized below:
\begin{itemize}
	\item It is assumed that VC remains stationary during each time slot\cite{b39}.
	\item We presume that a single vehicle can only handle one subtask at a time \cite{b21}. Consequently, if multiple subtasks are assigned to a vehicle, they must wait until resources become available\footnote{Upon having vehicles that can process multiple subtasks simultaneously, those vehicles can be modeled as multiple virtual vehicles with unlimited contact duration among them, where each of them can process one subtask at a time.}.		
	\item Due to the mobility and the limited contact durations among vehicles, this paper only focuses on one-hop data transmission between vehicles\cite{b38}. 
	\item Since the size of the feedback sent to the RSU is usually smaller than that of the original input data, the time it takes to transmit this feedback is neglected\cite{b30}. 
	
\end{itemize} 

\begin{table}[!t]
	\footnotesize
	\renewcommand{\arraystretch}{1.3}
	\caption{Summary Of The Notations Used In Our Problem Formulation}
	\label{table1}
	\centering
	\begin{tabular}{ll}
		\toprule[1pt]                                                                                                                 
		$\mathcal{V}$          & \begin{tabular}[c]{@{}l@{}}Set of vehicles covered by RSU \end{tabular}                                 \\\toprule[0.5pt] 	
		$\mathcal{B}$ & \begin{tabular}[c]{@{}l@{}}Set of subtasks contained by DAG\end{tabular} \\	\toprule[0.5pt]
		$\mathcal{E}$ & \begin{tabular}[c]{@{}l@{}}Set of edges associated with subtasks in $\mathcal{B}$ \end{tabular} \\\toprule[0.5pt]
		$\mathcal{P}_{i}$ & \begin{tabular}[c]{@{}l@{}}Set of immediate predecessors of subtask $b_i$ \end{tabular} \\\toprule[0.5pt]
		$\mathcal{S}_{i}$ & \begin{tabular}[c]{@{}l@{}}Set of immediate successors of subtask $b_i$ \end{tabular} \\\toprule[0.5pt]
		$\mathcal{B}_m$          & \begin{tabular}[c]{@{}l@{}}Set of subtasks executed on vehicle $v_m$ \end{tabular}                                 \\\toprule[0.5pt] 	
		$f_{m}$             & \begin{tabular}[c]{@{}l@{}}Computation capability of vehicle $v_m$\end{tabular}            \\\toprule[0.5pt]
		$d_{m,n}(t)$              & \begin{tabular}[c]{@{}l@{}}Euclidean distance between vehicles $v_m$ and $v_n$ at slot $t$
		\end{tabular}                                 \\\toprule[0.5pt]
		$u_{i}$              & \begin{tabular}[c]{@{}l@{}}Computation workload of subtask $b_i$
		\end{tabular}                                 \\\toprule[0.5pt]
		$c_{i,j}$         & Data transmission size between subtasks $b_i$ and $b_j$                                                                           \\\toprule[0.5pt]
		$AT_{m}$           & \begin{tabular}[c]{@{}l@{}}Arrival time of vehicle $v_m$ at VC \end{tabular}                                         \\\toprule[0.5pt]
		$DT_{m}$           & \begin{tabular}[c]{@{}l@{}}Departure time of vehicle $v_m$ from VC \end{tabular}                                         \\\toprule[0.5pt]
		$TT_{i,m;j,n}$           & \begin{tabular}[c]{@{}l@{}}Data transmission time between subtasks $b_i$ and $b_j$ when \\they are processed on vehicle $v_m$, $v_n$, respectively \end{tabular}                                         \\\toprule[0.5pt]
		$EST_{i,m}$           & \begin{tabular}[c]{@{}l@{}}Earliest start time of processing subtask $b_i$ on vehicle $v_m$\end{tabular}                                         \\\toprule[0.5pt]
		$AVT_{i,m}$         & \begin{tabular}[c]{@{}l@{}}Available time of vehicle $v_m$ when it is ready to process\\ subtask $b_i$ \end{tabular}                                         \\\toprule[0.5pt]
		$AFT_{i}$         & \begin{tabular}[c]{@{}l@{}}Actual finish time of subtask $b_i$ when it is practically \\ processed via a specific vehicle \end{tabular}                                         \\\toprule[0.5pt]
		$EFT_{i,m}$         & \begin{tabular}[c]{@{}l@{}}Earliest finish time of processing subtask $b_i$ on  $v_m$\end{tabular}                                         \\\toprule[0.5pt]
		$RT_{i}$         & \begin{tabular}[c]{@{}l@{}}Ready time of subtask $b_i$ when all of its \\ predecessors have been completed \end{tabular}                                         \\\toprule[0.5pt]
		$\xi_{i,m}$       & \begin{tabular}[c]{@{}l@{}}Binary variable indicating whether subtask $b_i$ is allocated\\ to vehicle $v_m$
		\end{tabular}     \\
		\toprule[1pt]                                                                                                       
	\end{tabular}
	\vspace{-0.5cm}
\end{table}
\subsection{DAG Task Model}
Without loss of generality, we index the task owner as $v_{1}$ with a computation-intensive DAG task, which is represented by a graph $\mathcal{G}=(\mathcal{B},\mathcal{E})$. In graph $\mathcal{G}$,  $\mathcal{B}=\{b_i \mid 1 \leq i \leq |\mathcal{B}|\}$ denotes the set of subtasks, and $\mathcal{E}$ denotes the set of directed edges, where  $e_{i,j} \in \mathcal{E}$ indicates that subtask $b_i$ has to be completed before the execution of subtask $b_j$. To better capture the sequential execution nature of DAG tasks, we further define the set of immediate \emph{predecessors} of each subtask $b_i$ as $\mathcal{P}_{i}=\{b_j \mid b_j \in \mathcal{B}, e_{j,i} \in \mathcal{E}\}$. Similarly, we define the set of immediate \emph{successors} of each subtask $b_i$ as $\mathcal{S}_{i}$. For example, in Fig. \ref{fig_1}, we have $\mathcal{B}=\{b_1,b_2,b_3,b_4\}$, $\mathcal{E}=\{e_{1,2},e_{1,3},e_{2,4},e_{3,4}\}$, $\mathcal{P}_{4}=\{b_2,b_3\}$, and $\mathcal{S}_{1}=\{b_2,b_3\}$. Furthermore, to make our analysis tractable, we introduce a virtual subtask to the DAG task topology denoted by $b_{0}$, which is connected to subtask(s) with no immediate predecessors as shown in Fig. \ref{fig_2}. 
\vspace{-0.1cm}
\subsection{Vehicle Mobility Model}
We assume that each vehicle $v_m$ is driving at a random and constant speed $g_{m}$ (meters per second). Since the speeds of vehicles are non-negative, we adopt a truncated Gaussian distribution\cite{b32} to capture them. Specifically, for any value of speed $g$, the probability density function of the truncated Gaussian distribution is defined as
\begin{flalign}
	\widehat{F}(g)=\frac{2F\left(g\right)}{\Phi\left(\frac{g_\mathsf{max }-\mu_g}{\sigma_g\sqrt{2}}\right)-\Phi\left(\frac{g_\mathsf{min }-\mu_g}{\sigma_g\sqrt{2}}\right)},\label {eq_1}
\end{flalign}
where $\Phi(x)=\frac{2}{\sqrt{2 \pi}}\int_{0}^{x} e^{-t^2} dt$ is the Gaussian error function, and $g_\mathsf{max}$ and $g_\mathsf{min}$ are defined as the maximum and minimum speed of vehicles, respectively. In (\ref{eq_1}), $	F\left(g\right)$ is the probability density function of a Gaussian distribution which is given by
\begin{flalign}
	F\left(g\right)=\frac{1}{\sigma_g\sqrt{2 \pi}} \exp \left(\frac{-(g-\mu_g)^2}{{2}\sigma_g^2}\right), \label {eq_2}
\end{flalign}
where $\mu_g$ is the average speed of all vehicles, and $\sigma_g$ is the standard deviation. 


Considering resource provisioning for DAG subtasks is conducted by vehicles that are located in the VC (i.e., within the coverage of the RSU), we utilize the notion of the \emph{dwell time} to characterize vehicles' mobility. Specifically, considering that a contact event (i.e., V2V link formation) can happen between two vehicles as long as they have not left the VC, we define the dwell time of vehicle $v_m$ in the VC as interval $[AT_m, DT_m]$, where $AT_m$ and $DT_m$ represent the arrival and departure time of $v_m$ at and from the VC, respectively, between which vehicle $v_m$ is available to offer its computation resource. 

\subsection{Computation Offloading Model}
\textbf{Path Loss Model.} Let $(x_{m}(t),y_{m}(t))$ denote the 2D coordinates of each vehicle $v_m$ at time slot $t$, to consider the impact of dynamics of VCs on V2V links, we first adopt a dual-slope piecewise-linear model\cite{b29} to represent the propagation loss (in dB) between two vehicles $v_m$ and $v_n$, denoted by $PL\left(d_{m,n}(t)\right)$, as follows:
\begin{flalign}
	 PL\left(d_{m,n}(t)\right)= PL_{\mathsf{LoS}}\left(d_{m,n}(t)\right) + \beta, \ \forall v_{m},v_{n} \in \mathcal{V},\label {eq_3}
\end{flalign}
where $d_{m,n}(t)=\sqrt{(x_{m}(t)-x_{n}(t))^2+(y_{m}(t)-y_{n}(t))^2}$ (in meters) denotes the Euclidean distance between vehicles $v_{m}$ and $v_{n}$ at time slot $t$, and $\beta$ is an additional attenuation factor modeled according to a lognormal random variable with
mean $\mu_{\beta}=5+\text{max}(0,15\text{log}_{10}(d_{m,n}(t))-41)$ (in dB) and standard deviation $\sigma_{ \beta}=4.5$ (in dB). In (\ref{eq_3}) $PL_{\mathsf{LoS}}\left(d_{m,n}(t)\right)$ is the path loss of the light-of-sight (LoS) transmission between two vehicles, which is given by
\begin{flalign}
	&PL_{\mathsf{LoS}}\left(d_{m,n}(t)\right)=32.4+20\text{log}_{10}(d_{m,n}(t))\nonumber\\
	&+20\text{log}_{10}(F_c)+ \delta, \ \forall v_{m},v_{n} \in \mathcal{V}, \label {eq_4}
\end{flalign}
where $F_c$ is the center frequency (in GHz), and $\delta$ captures the effect of signal power fluctuations due to surrounding objects modeled by a lognormal random variable with standard deviation $\sigma_{\delta}=3$ (in dB).
	
We then introduce the notion of \textit{ready time} which enables us to develop  our scheduling methodology for DAG tasks by taking their sequential execution into account. 
\begin{myDef}
	\textbf{(Ready Time).} Ready time $RT_i$ indicates the time when all of the immediate predecessors of subtask $b_i$ are completed/finished, which corresponds to the starting time of data transmission between $b_j$ ($b_j \in \mathcal{P}_i$) and the vehicle that processes $b_i$:
	\begin{equation}
		RT_{i}=\max _{b_j \in {\mathcal{P}_i}}\{AFT_{j}\}, \ b_i \in \mathcal{B}, \label {eq_5}
	\end{equation}
	where $AFT_{j}$ is the actual finish time of subtask $b_j$ when it is practically executed on a vehicle.
\end{myDef}

\textbf{Transmission Model.} Combining (\ref{eq_3}) - (\ref{eq_5}), we let $TT_{i,m;j,n}$ denote the data transmission time associated with edge $e_{i,j}$ when subtasks $b_i$ and $b_j$ are allocated to vehicles $v_{m}$ and $v_{n}$, respectively, which can be calculated as
	\begin{flalign}
		TT_{i,m;j,n}=\left\{\begin{array}{l}
			c_{i,j}{ \Psi\left(PL\left(d_{m,n}(RT_j)\right)\right)}, \ m\neq n  \\
			0, \quad\quad\quad\quad\quad\quad\quad \ m = n
		\end{array}\right. \nonumber \\
		\forall b_i,b_j \in \mathcal{B}, \ e_{i,j} \in \mathcal{E}, \ v_{m},v_{n} \in \mathcal{V},\label {eq_6}
	\end{flalign}
	where $	c_{i,j}$ (in bits) is the transmission data size between subtasks $b_i$ and $b_j$ associated with directed edge $e_{i,j}$, and $\Psi(\cdot)$ is a monotone increasing function indicating that a higher value of path loss between vehicles $v_m$ and $v_n$ at time slot $RT_j$ (i.e., a worse V2V channel condition) leads to a longer transmission time (see Section V for a realization of $\Psi(\cdot)$).
	
\textbf{Computation Model.} To model the scheduling of DAG subtasks, let $EST_{i,m}$ and $EFT_{i,m}$ denote the earliest start time, and finish time of processing of subtask $b_i$ on vehicle $v_{m}$, respectively. We assume that virtual subtask $b_0$ is processed on the task owner (i.e., $v_1$) and its computation workload is zero, we thus have $EST_{0,1}=EFT_{0,1}=0$. 
	
Subsequently, for each subtask $b_i \in \mathcal{B}, i \neq 0$, the values of $EST_{i,m}$ and $EFT_{i,m}$ can be calculated recursively as follows: 
\begin{flalign}	
	&EST_{i,m}=\max \left\{\! AVT_{i,m}, \overbrace{RT_{i}+\max _{b_j \in \mathcal{P}_{i}}\{ TT_{j,n;i,m}\}}^{\text{(I)}} \right\}, \nonumber \\
	&\quad \quad\quad\quad\quad\quad\quad\quad\quad\quad\quad\quad\quad\ \forall b_i \in \mathcal{B}, \ v_m \in \mathcal{V},
		\label {eq_7}	
\end{flalign}
where $AVT_{i,m}$ denotes the available time when vehicle $v_m$ completes its latest assigned subtask and term (I) indicates the earliest arrival time of the required data for processing subtask $b_i$ at vehicle $v_m$. 

Furthermore, we consider heterogeneous computation capabilities across vehicles, where for each vehicle $v_m$, its computation capability is denoted by $f_m$ (in CPU cycles per second). As a result, the earliest finish time of processing subtask $b_i$ on vehicle $v_m$ is given by 
\begin{flalign}	
	EFT_{i,m}\!=\! EST_{i,m}\!+\!\frac{u_i}{f_m}, \ \forall b_i \in \mathcal{B},  v_m \in \mathcal{V}, \label {eq_8}
\end{flalign}
where $u_i$ denotes the computation workload (in CPU cycles) of subtask $b_i$.  

\begin{figure}[!t]
	\centering
	\includegraphics[width=3.5in]{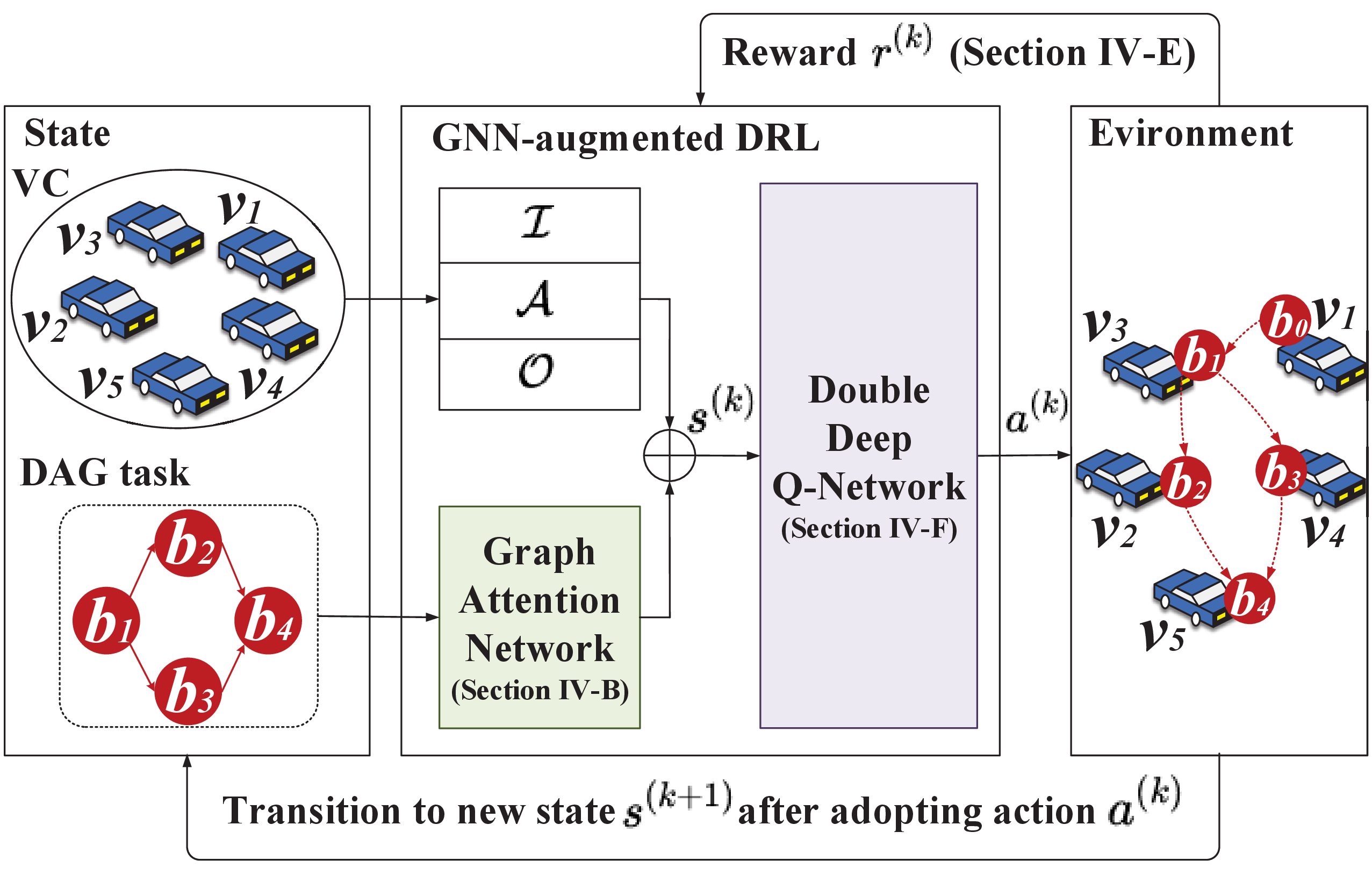}
	\caption{A schematic of our GNN-augmented DRL scheme for VC-assisted DAG task scheduling. }
	\label{fig_3}
	\vspace{-0.5cm}
\end{figure}

\subsection{Optimization Formulation}
We capture the subtask-to-vehicle allocations through a set of binary indicators $\mathcal{I}=\{\xi_{i,m} \mid 0 \leq i \leq |\mathcal{B}|, \ 1 \leq m \leq |\mathcal{V}|\}$, where $\xi_{i,m}=1$ denotes that subtask $b_i$ is allocated to vehicle $v_m$, and $\xi_{i,m}=0$ otherwise. Aiming to minimize the overall DAG task completion time, we formulate DAG task scheduling over the VC as the following mixed integer programming (MIP):
\begin{flalign}
	&\quad\quad\quad\quad\quad \min_{\mathcal{I}} \ \max _{b_i \in \mathcal{B}, v_m \in \mathcal{V}}\{EFT_{i,m} \xi_{i,m}\}, \label {eq_9}  \\        
	&\text{s.t.} \ (\ref{eq_7}), (\ref{eq_8}),\nonumber\\ 
	&\sum_{v_m \in \mathcal{V}} \xi_{i, m}=1, \ b_i \in \mathcal{B}, \tag{C1}\\
	&\xi_{i,m} \in\{0,1\}, \ b_i \in \mathcal{B}, \ v_m \in \mathcal{V},      \tag{C2}\\
	&\!\!\bigcap _{b_i \in \mathcal{B}_{m}}\!\!\left[EST_{i, m},EFT_{i, m}\right]=  \varnothing, \ v_m \in \mathcal{V},    \tag{C3}\\
	&EST_{i,m} \geq EFT_{j,n},\ b_i \in \mathcal{B}, \ b_j\in \mathcal{P}_{i},\ v_m,v_n \in \mathcal{V}\tag{C4}, \\
	&\left[EST_{i, m},EFT_{i, m}\right] \subset \left[AT_m,DT_m\right],\ b_i \in \mathcal{B}, \ v_m \in \mathcal{V}. \hspace{-1em}\tag{C5}
\end{flalign}
In (\ref{eq_9}), the objective function captures the sequential execution of DAG subtasks, where the maximum finish time of all subtasks indicates the overall DAG task completion time. Also, constraint (C1) guarantees that each subtask is allocated to only one vehicle, while (C2) restricts the value of the allocation indicator $\xi_{i,m}$ to be binary. Constraint (C3) ensures that a vehicle can only process one subtask at a time, where $\mathcal{B}_m=\{b_i \mid \xi_{i,m}=1,\ 0 \leq i \leq |\mathcal{B}|\}$ denotes the set of subtasks processed on vehicle $v_m$. Constraint (C4) indicates that the processing of a subtask can not start until all of its predecessors are completed, (C5) guarantees the availability of computation resources of vehicles with respect to the vehicles' dwell times: the earliest start time and earliest finish time of executing each subtask $b_i$ on vehicle $v_m$ should between the arrival time and departure time of vehicle $v_m$ in a  VC.

It is known that MIP formulations similar to what we have in (\ref{eq_9}) are NP-hard\cite{b36}. Also, considering the sequential execution of DAG subtasks (i.e. different subtasks may be executed at different time slots), we need the prior knowledge of the V2V path loss and availability of vehicles' computation resources, obtaining of which is cumbersome in practice. As a result, to tackle these challenges, we propose a GNN-augmented DRL scheme, named GA-DRL, to efficiently find near-optimal solutions for (\ref{eq_9}).
\begin{table}[!t]
	\footnotesize
	\renewcommand{\arraystretch}{1.3}
	\caption{Summary Of The Notations Used In GA-DRL Design}
	\label{table2}
	\centering
	\begin{tabular}{ll}
		\toprule[1pt]                                                                                                                 
		$h^{(0)}_i$          & \begin{tabular}[c]{@{}l@{}}Raw feature of subtask $b_i$ \end{tabular}                                 \\\toprule[0.5pt] 
		$\mathcal{N}_i$          & \begin{tabular}[c]{@{}l@{}}Neighbor set of subtask $b_i$ \end{tabular}                                 \\\toprule[0.5pt]	
		$\mathcal{N}^{-1}_i$          & \begin{tabular}[c]{@{}l@{}}Inverse neighbor set of subtask $b_i$ \end{tabular}                                 \\\toprule[0.5pt]	
		$h_i^{(\ell)}$ & \begin{tabular}[c]{@{}l@{}}Result feature of subtask $b_i$ at iteration $\ell$\end{tabular} \\	\toprule[0.5pt]
		$\alpha_{i,j}^{(\ell)}$ & \begin{tabular}[c]{@{}l@{}}Attention coefficient between subtasks $b_i$ and $b_j$ at iteration $\ell$ \end{tabular} \\\toprule[0.5pt]
		$W^{(\ell)}$ & \begin{tabular}[c]{@{}l@{}}Trainable weight matrix of GAT at iteration $\ell$ \end{tabular} \\\toprule[0.5pt]
		$A^{(\ell)}$ & \begin{tabular}[c]{@{}l@{}}Trainable feedforward neural network at iteration $\ell$ \end{tabular} \\\toprule[0.5pt]
		$\mathcal{L}^{\mathsf{rank}}$ & \begin{tabular}[c]{@{}l@{}}Subtask scheduling priority list \end{tabular}  \\
		\toprule[1pt]                                                                                                        
	\end{tabular}
	\vspace{-0.5cm}
\end{table}

\section{GNN-Augmented DRL (GA-DRL) for DAG Task Scheduling over Dynamic VCs}
In this section, we first provide an overview of our GA-DRL methodology and the challenges we aim to address. We then tailor a GAT module for extracting features of subtasks. Subsequently, the VC-assisted DAG task scheduling is modeled as an MDP consisting of the state space, action space, and reward. Finally, we utilize a DDQN architecture to tackle (\ref{eq_9}) and discuss its training procedure.
\subsection{GA-DRL Overview and Challenges}
\subsubsection{GA-DRL overview} Our method takes a different approach from traditional DRL methods developed for task scheduling \cite{b16},\cite{b18},\cite{b25}, which only consider predetermined states, such as computation workload, data size, and number of subtask predecessors/successors. Instead, we propose a GNN-augmented DRL approach that automatically learns distinctive subtask features and creates assignments between subtasks and vehicles. In particular, as shown in Fig. \ref{fig_3}, the features of subtasks are acquired through a GAT module, rather than being predetermined.

Our GA-DRL conducts subtask-to-vehicle allocations through a sequence of decision steps. At each decision step $k$, the DRL agent functioning at the RSU diligently collects relevant data on the system \textit{state} $s^{(k)}$, which includes the extracted features of current subtask obtained by GAT, as well as the parameters of the vehicles describing their dynamics and heterogeneity. DRL agent then feeds state $s^{(k)}$ to a DDQN. The objective of DDQN is to effectively assign subtasks to vehicles by determining the best course of \textit{action} $a^{(k)}$. To this end, DDQN evaluates the value of each state-action combinations, and conducts a subtask-to-vehicle allocation $a^{(k)}$, which moves the system to the next state $s^{(k+1)}$. Finally, the DRL agent receives a \textit{reward} $r^{(k)}$, that aids in the training of a deep learning model. This, in turn, enhances the agent's ability to take better actions over time.
\subsubsection{Main challenges} When applying GA-DRL to the VC-assisted DAG task scheduling, there are two main challenges that need to be tackled.

\textbf{(1) Feasibility of allocation decisions.} Unlike static computing environments that have stable, fully-connected computing servers\cite{b3},\cite{b4},\cite{b6}, the dynamics of VC's resources can greatly affect the execution of DAG subtasks. This is captured by constraint (C5) in (\ref{eq_9}), satisfying of which guarantees the time-interval of processing subtask $b_i$ on vehicle $v_m$ to be within the dwell time of $v_m$. Ensuring that subtask-to-vehicle allocation decisions are feasible (specifically, meeting constraint (C5)) can be difficult because neural networks typically lack a module to filter out infeasible actions.

\textbf{(2) Generalizability of designed GNN.} Efficient inductive learning is a key feature of GAT\cite{b31}, which makes it suitable for using with previously unseen graph topologies. However, it can be difficult to ensure that the GNN model is applicable to various DAG tasks, as each task has its own unique topology and interdependence among subtasks. To overcome this challenge, we must carefully encode the information of each DAG task's topology to achieve meaningful results when combined with our later developed GAT.

Table \ref{table2} summarizes the major notations used in this section.
\begin{figure}[!t]
	\centering
	\includegraphics[width=3.5in]{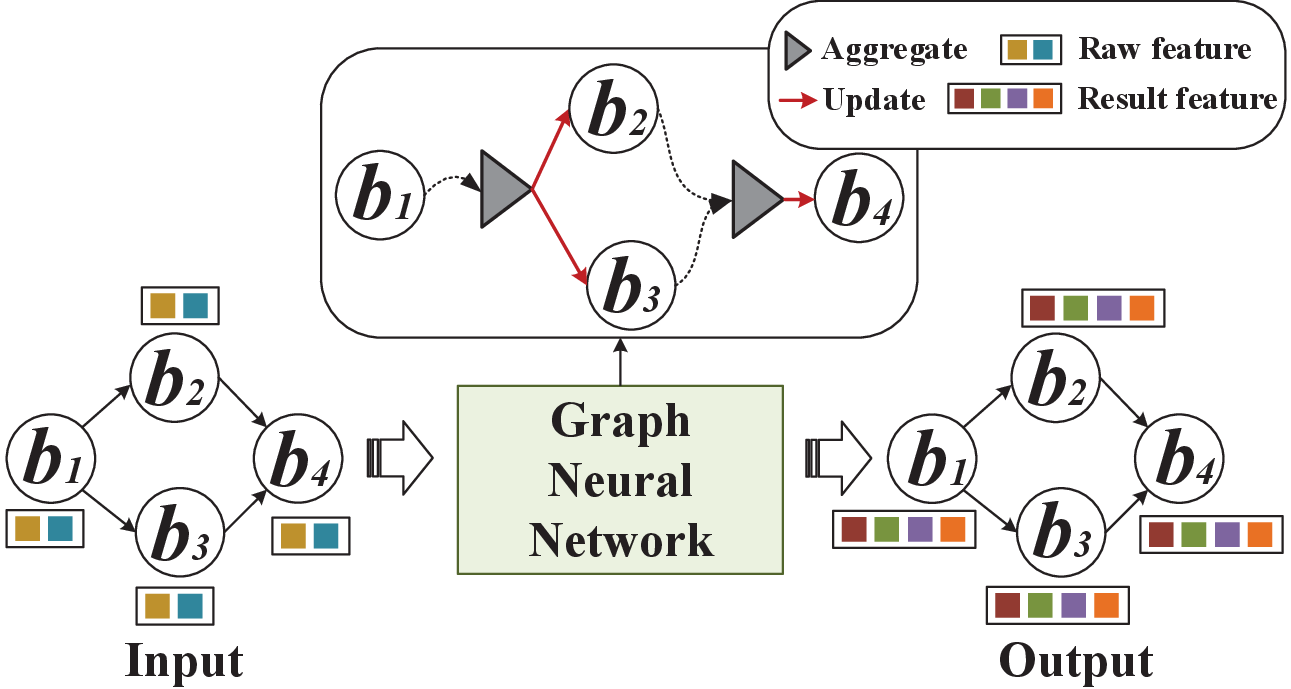}
	\caption{A schematic of architecture of GNN for extracting the features of the DAG task depicted in Fig. \ref{fig_1}. The colored squares in the diagram correspond to different features.}
	\label{fig_4}
	\vspace{-0.5cm}
\end{figure}
\vspace{-0.5cm}
\subsection{Graph Neural Network}
In this subsection, we explain the structure of GNNs and how we use a multi-head GAT to extract distinctive features of subtasks. Our GAT incorporates a two-way aggregation method that considers the topological information of both predecessors and successors of each subtask. To further enhance the adaptability of our GAT to new DAG tasks, we utilize a ranking-based sampling technique.

\subsubsection{Architecture of GNNs} The architecture of a GNN is depicted in Fig. \ref{fig_4}, where a GNN takes raw features of all subtasks as the input and subsequently generates result features containing corresponding topological information of the DAG task. Specifically, GNN utilizes an $\textbf{Aggregate}$ function to accumulate the topological information passed by the neighbors of each subtask. The accumulated information is then modified through a nonlinear $\textbf{Update}$ function. This procedure is repeated $L$ times to create the result feature for each subtask.

\textbf{Raw feature of each subtask.} Similar to conventional DRL methods \cite{b20,b21,b22,b23,b24}, which rely on human-selected information to define DAG subtasks, we also define the raw feature\footnote{Super-index $0$ is used to capture that these are initial features of the subtask, which are later processed and enhanced through GNN.} of each subtask $b_i$ as
\begin{flalign}
	h^{(0)}_{i}=\{u_i, \overline{c_{i}}, |\mathcal{P}_i|, |\mathcal{S}_i|\}, \ b_i \in \mathcal{B},\ b_j \in \mathcal{S}_i,\label{eq_10}		
\end{flalign}
where $u_i$ is the computation workload of subtask $b_i$, and $\overline{c_{i}} = 
\sum_{b_j \in \mathcal{S}_i} \frac{c_{i, j}}{|\mathcal{S}_i|}$ indicates the average transmission data size associated with edges $e_{i,j},\ b_j \in \mathcal{S}_i$. Also, $|\mathcal{P}_i|$ and $|\mathcal{S}_i|$ represent the number of predecessors and successors of subtask $b_i$, respectively.

\textbf{Neighbor set of each subtask.} Considering that DAG subtasks are executed sequentially, we define $\mathcal{N}_i$ as the neighbor set of each subtask $b_i$ which includes all of its predecessors as well as $b_i$ itself; mathematically 
\begin{flalign}
	\mathcal{N}_{i}= \{b_j \mid e_{j,i}\in \mathcal{E}\}\cup\{b_i\}. \label{eq_11}		
\end{flalign}

Through an iterative process involving the use of $\textbf{Update}$ and $\textbf{Aggregate}$ functions, GNN obtains the result feature of each subtask $b_i$. Mathematically, at each iteration $\ell$, we have
\begin{flalign}
	h^{(\ell+1)}_{i}= \textbf{Update}(\textbf{Aggregate}(\{h^{(\ell)}_{j}|b_j \in \mathcal{N}_i\})), \label{eq_12}		
\end{flalign}
where $h^{(\ell+1)}_{i}$ denotes the result feature of subtask $b_i$ at iteration $\ell$. Through $L$ iterations, the GNN derives the final result feature for each subtask, denoted by $h_i^{(L)}$. This feature incorporates both the raw feature of each subtask (i.e., at $\ell = 0; h^{(0)}_i$), as well as the topological information from neighboring subtasks (i.e., $\mathcal{N}_i$) within the DAG task.

Hereafter, we detail the $\textbf{Aggregate}$ and the $\textbf{Update}$ functions designed to extract features of DAG subtasks.

\subsubsection{Multi-head GAT}
Considering that the subtasks involved in $\mathcal{N}_i$ have different computation workloads, transmission data sizes and interdependencies, we employ an attention mechanism, which is inspired by \cite{b31} to assign diverse weights to subtasks with the aim of enhancing information of key subtasks. Specifically, at each iteration $\ell$, we define an attention-based aggregation function called $\textbf{Aggregate}^{\mathsf{at}}$ as
\begin{flalign}
	\hspace{-0.5em}\textbf{Aggregate}^{\mathsf{at}}(\{ h^{(\ell)}_{j}|b_j \in \mathcal{N}_i\})=\!\!\!\sum_{b_j \in \mathcal{N}_i}\alpha^{(\ell)}_{i,j}W^{(\ell)}h^{(\ell)}_{j}, \hspace{-0.5em}\label{eq_13}		
\end{flalign} 
where $W^{(\ell)}$ is a trainable weight matrix at iteration $\ell$, and $\alpha^{(\ell)}_{i,j}$ is a normalized attention coefficient at iteration $\ell$, which measures the relative importance of subtask $b_j$ to subtask $b_i$ as follows:
\begin{flalign}
	\alpha^{(\ell)}_{i,j}= \frac{\exp \left(A^{(\ell)} [ W^{(\ell)} h^{(\ell)}_i || W^{(\ell)}h^{(\ell)}_j] \right)}{\sum\limits_{b_{j^{\prime}} \in \mathcal{N}_i} \exp \left(A^{(\ell)}[W^{(\ell)}h^{(\ell)}_i || W^{(\ell)}h^{(\ell)}_{j^{\prime}}] \right)}.\label{eq_14}		
\end{flalign} 
In (\ref{eq_14}), $A^{(\ell)}$ is a trainable vectors at iteration $\ell$, and $\cdot||\cdot$ denotes the vector concatenation.

Further, in order to enhance the effectiveness of GAT's learning process, we propose to use a multi-head GAT, where different attention heads learn to give more relevant weights to different subtasks. Let $Z$ denote the total number of heads. Each attention head, denoted by $z$ will individually aggregate topological information of subtasks, in conjunction with other attention modules. The multi-head attention-based aggregation function called $\textbf{Aggregate}^\mathsf{mat}$ can be then formulated as 
\begin{flalign}
	& \textbf{Aggregate}^\mathsf{mat}(\{h^{(\ell)}_{j}|b_j \in \mathcal{N}_i\})\nonumber\\
	&= \frac{1}{Z} \sum_{z=1}^Z \left(\sum_{b_j \in \mathcal{N}_i}\!\! \alpha^{(\ell){(z)}}_{i,j}W^{(\ell)(z)}h^{(\ell)}_{j}\right), \hspace{-0.5em}\label{eq_15}	
\end{flalign} 
where iteration index $\ell$ and head index $z$ are both used as superscripts hereafter. 

To better suit our problem, we aim to modify the $\textbf{Aggregate}^{\mathsf{mat}}$ function defined in (\ref{eq_15}) through developing a two-way aggregation for the multi-head GAT. This approach takes into consideration the predecessors and successors of each subtask, which helps to aggregate topological information in a more effective manner.

\subsubsection{Two-way aggregation}
To execute DAG subtasks, capturing the conditions of predecessors and successors  of each subtasks are equally important. As a result, we develop a two-way aggregation approach that utilizes two different types of attention heads. This approach involves using the inverse neighbor set  $\mathcal{N}_i^{-1}$ of each subtask $b_i$, which includes all of its successors and $b_i$ itself 
\begin{flalign}
	\mathcal{N}^{-1}_i=\{b_j \mid e_{i,j} \in \mathcal{E}\} \cup \{b_i\}. \label{eq_16}		
\end{flalign}

At each iteration $\ell$, half of the attention heads from $Z$ are then allocated to collect topological information from the neighboring subtasks, while the remaining half is utilized to gather topological information from the inverse neighbor set, which leads to the modification of (\ref{eq_15}) to
\vspace{-0.3em}
\begin{flalign}
	&\textbf{Aggregate}^\mathsf{mat}(h^{(\ell)}_{j})=\frac{1}{Z}\Bigg[\sum_{z=1}^{Z/2}\left(\sum_{b_j \in \mathcal{N}_i} \alpha^{(\ell)(z)}_{i,j}W^{(\ell)(z)}h^{(\ell)}_{j}\right) \nonumber \\
	&\quad\quad\quad\quad\quad +\sum_{z=Z/2}^{Z}\left(\sum_{b_j \in \mathcal{N}^{-1}_i} \alpha^{(\ell)(z)}_{i,j}W^{(\ell)(z)}h^{(\ell)}_{j}\right)\Bigg]. \label{eq_17}		
\end{flalign}

We then aim to further modify the $\textbf{Aggregate}^{\mathsf{mat}}$ function defined in (\ref{eq_17}). This makes out GAT module different from other existing GNNs \cite{b34,b35,b36,b37}: we do not consider all the neighbors of a given subtask to accumulate topological information. Instead, we opt for a weighted sample of neighbors, based on their scheduling priority. This approach allows our GAT to be more generalizable to unseen DAG task topologies. We next describe this approach.

\subsubsection{Ranking-based sampling} We first devise an approach to  prioritize scheduling of subtasks based on their ranking value. By employing a recursive method, we determine the ranking value of each subtask $b_i$ labeled as $rank_i$ as follows:
\begin{flalign}
	rank_i = \max_{b_j \in \mathcal{P}_i} \{{rank_j} +\overline{u_{j}} + \overline{c_{j,i}}\}, \ b_i \in \mathcal{B},\label {eq_18}	
\end{flalign}
where $\overline{u_{j}}$ is the average execution cost of subtask $b_j, b_j \in \mathcal{P}_i$, which is given by 
\begin{flalign}
	\overline{u_{j}}=\frac{\sum_{m=1}^{\left|\mathcal{V}\right|}u_j / f_m}{\left|\mathcal{V}\right|}, \ v_m \in \mathcal{V}, \label {eq_19}	
\end{flalign} 
and $\overline{c_{j,i}}$ denotes the average transmission cost associated with edge $e_{j,i}, b_j \in \mathcal{P}_i$ at the beginning (i.e., at time slot $0$), which is given by
\begin{flalign}
	\overline{c_{j,i}}=\frac{\sum_{m=1}^{\left|\mathcal{V}\right|}\sum_{n=1}^{\left|\mathcal{V}\right|}c_{j,i}\Psi\left(PL\left(d_{m,n}\left(0 \right)\right)\right)}{\left|\mathcal{V}\right|^{2}}. \label {eq_20}	
\end{flalign}

Assuming $rank_0=0$ for virtual subtask $b_0$, we maintain a subtask scheduling priority list $\mathcal{L}^{\mathsf{rank}}$ as
\begin{flalign}
	\mathcal{L}^{\mathsf{rank}}=\{b_i \succ b_j \mid b_i,b_j \in \mathcal{B}, \ rank_i <rank_j \}, \label {eq_21}	
\end{flalign}
where the preference relation $b_i \succ b_j$ indicates that subtask $b_i$ has a higher scheduling priority compared with subtask $b_j$ due to a lower value of $rank_i$\footnote{Our current ranking method for DAG subtasks relies on heuristics, which may limit the GNN-augmented DRL algorithm's ability. We plan to address this issue by exploring alternative methods for determining the scheduling priority of DAG subtasks using DRL in the future.}.

Finally, we define $\mathcal{N}^{\mathsf{rank}}_{i}$ as a ranking-based neighbor set of subtask $b_i$ which contains the subtasks sampled from $\mathcal{N}_{i}$. The sampling probability/weight of subtask $b_j$ from $\mathcal{N}_{i}$ to be included in $\mathcal{N}_i^{\mathsf{rank}}$, denoted by $p_j $, is calculated as
\begin{flalign}
	p_j=\frac{\exp\left(rank_j\right)}{\sum_{b_{j^{\prime}} \in \mathcal{P}_i}\exp\left(rank_{j^{\prime}}\right)}. \label {eq_22}	
\end{flalign}
This weighted subtask sampling method leads to improving the generalizability of our method by intentionally  losing topological information passed by the subtasks which are not sampled, which makes our GAT model less sensitive to the topological variations in DAG tasks. This resembles the dropout\cite{b19} mechanism widely leveraged in training deep neural network. Note that subtask sampling is done with replacement if the sample size is larger than the size of $\mathcal{N}_{i}$.

\textbf{Aggregate function.} By integrating aforementioned methodologies, our designed $\textbf{Aggregate}^\mathsf{mat}$ function not only enables information enhancement of key subtasks by considering a two-way multi-head attention-based aggregation, but also improves generalizability by considering a ranking-based sampling; mathematically
\begin{flalign}
	&\textbf{Aggregate}^\mathsf{mat}(h^{(\ell)}_{j})=\frac{1}{Z}\Bigg[\sum_{z=1}^{Z/2}\left(\sum_{b_j \in \mathcal{N}^\mathsf{rank}_i} \alpha^{(\ell)(z)}_{i,j}W^{(\ell)(z)}h^{(\ell)}_{j}\right) \nonumber \\
	&\quad\quad\quad\quad +\sum_{z=Z/2}^{Z}\left(\sum_{b_j \in \mathcal{N}^{-\mathsf{rank}}_i} \alpha^{(\ell)(z)}_{i,j}W^{(\ell)(z)}h^{(\ell)}_{j}\right)\Bigg], \label{eq_23}		
\end{flalign}
where $\mathcal{N}^{-\mathsf{rank}}_i$ is the inverse ranking-based neighbor set of subtask $b_i$ sampled from the $\mathcal{N}^{-1}_{i}$ using a similar sampling method described in (\ref{eq_22}).

\textbf{Update function.} After receiving aggregated topological information in (\ref{eq_23}), we apply the exponential linear unit activation (ELU)\cite{b48} in the $\textbf{Update}$ function. Finally, combining the aforementioned $\textbf{Aggregate}$ and $\textbf{Update}$ functions, we can express (\ref{eq_12}) as
\begin{flalign}
	h^{(\ell+1)}_{i}=\textbf{ELU}\Bigg(\frac{1}{Z}\Bigg[\sum_{z=1}^{Z/2}\left(\sum_{b_j \in \mathcal{N}^{\mathsf{rank}}_i} \alpha^{(\ell)(z)}_{i,j}W^{(\ell)(z)}h^{(\ell)}_{j}\right) \nonumber \\
	+\sum_{z=Z/2}^{Z}\left(\sum_{b_j \in \mathcal{N}^{-\mathsf{rank}}_i} \alpha^{(\ell)(z)}_{i,j}W^{(\ell)(z)}h^{(\ell)}_{j}\right)\Bigg]\Bigg). \label{eq_24}		
\end{flalign}
In our experiments, we found that our approach could achieve high performance with $L=2,Z=4$, where $W^{(1)} \in \mathbb{R}^{4\times16}$, $A^{(1)} \in \mathbb{R}^{32\times1}$, and $W^{(2)} \in \mathbb{R}^{16\times32}$, $A^{(2)} \in \mathbb{R}^{64\times1}$. 
\begin{figure}[!t]
	\centering
	\includegraphics[width=3.5in]{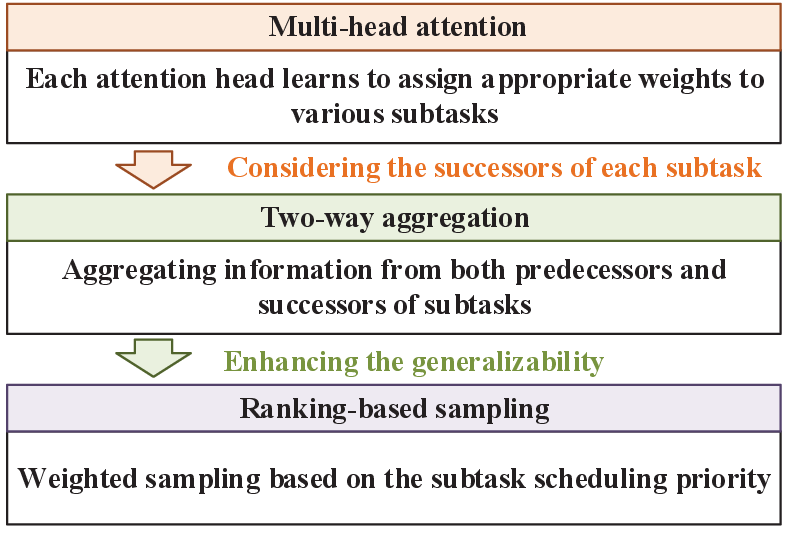}
	\caption{A flow chart of the relationships between different components of our methodology in Section IV-B.}
	\label{fig_5}
	\vspace{-0.5cm}
\end{figure}

A flow chart of the relationships between the components developed for $\textbf{Aggregate}$ function is shown in Fig. \ref{fig_5}. Also, \textbf{Algorithm} \ref{algorithm_1} details the corresponding procedure of our GAT module with computation complexity $\mathcal{O}(|\mathcal{B}|LZ)$, where we assume a set of learned parameters (i.e., $W^{(\ell)(z)}$ and $A^{(\ell)(z)}$). These parameters are later optimized in conjunction with DDQN parameters. We next formulate DAG task scheduling as an MDP with state, action, and reward representation.

\begin{algorithm} [!t]
	\small
	\centering 
	\caption {Ranking-based two-way multi-head GAT}\label{algorithm_1}
	\begin{algorithmic}[1]
		\STATE \textbf{Input:} Graph $\mathcal{G}=(\mathcal{B},\mathcal{E})$; raw features of subtasks $\{h^{(0)}_{i}, b_i\in \mathcal{B}\}$; iteration number $L$; head number $Z$; weight {\small $\{W^{(\ell)(z)}, \forall \ell \in \{1,\cdots,L\}, \forall z \in \{1,\cdots,Z\}\}$}; network {\small $\{A^{(\ell)(z)}, \forall \ell \in \{1,\cdots,L\}, \forall z \in \{1,\cdots,Z\}\}$}
		\STATE \textbf{Output:} Result features $h^{(L)}_i$ for all subtasks $b_i \in \mathcal{B}$
		\STATE Obtain $\mathcal{L}^{\mathsf{rank}}$ by (\ref{eq_18})- (\ref{eq_22})
		\FOR {$i=1\cdots |\mathcal{B}|$}
		\STATE \textbf{//Ranking-based sampling}\\ Weighted sample a fixed-size set $\mathcal{N}^{\mathsf{rank}}_i$ of neighbors in $ \mathcal{N}_i$ according to (\ref{eq_11}) and (\ref{eq_22})
		\FOR {$\ell=1\cdots L$}
		\STATE \textbf{//Multi-head attention}
		\FOR {$z=1\cdots Z$}
		\FOR {$j=1\cdots \mathcal{N}^{\mathsf{rank}}_i$}
		\STATE Calculate $\alpha^{(\ell)(z)}_{i,j}$ by (\ref{eq_14}) 
		\ENDFOR 
		\STATE \textbf{//Two-way aggregation} 
		\STATE Weighted sample a fixed-size set $ \mathcal{N}^{-\mathsf{rank}}_i$ of neighbors in $\mathcal{N}^{-1}_i$
		\STATE Considering $Z$ attention heads, use (\ref{eq_23}) to aggregate the massages passed by neighbors in $\mathcal{N}^{\mathsf{rank}}_i$ as well as $\mathcal{N}^{-\mathsf{rank}}_i$
		\STATE Updating the accumulative features using (\ref{eq_24})
		\ENDFOR
		\ENDFOR
		\ENDFOR		
	\end{algorithmic}
\end{algorithm}

\subsection{State Representation}
The result feature of each subtask $b_{i}$, denoted by $h^{(L)}_{i}$, is generated through consecutive $L$ iterations in (\ref{eq_24}). We assume subtasks to vehicles assignments through a series of decision steps indexed by $k$, at each decision step $k$, let subtask $b_{\tau(k)}$ be the \textit{current subtask} waiting to be allocated to a vehicle, where $\tau(k)$ indicates the subtask's index at position $k$ in
$\mathcal{L}^{\mathsf{rank}}$. We define the system state $s^{(k)}$ as follows:
\begin{flalign}
	s^{(k)}=\Big\{h^{(L)}_{\tau{(k)}}, \mathcal{I}^{(k-1)}, \mathcal{A}^{(k)}, \mathcal{O}^{(k)}\Big\}, \label{eq_25}		
\end{flalign}
where $\mathcal{I}^{(k-1)}$ denotes subtask-to-vehicles allocation decisions for the subtasks located before current subtask $b_{\tau(k)}$ in $\mathcal{L}^{\mathsf{rank}}$, and $\mathcal{A}^{(k)}= \{avail_1, avail_2, \cdots, avail_{|\mathcal{V}|}\}$ is the availability indicator set at the instant of decision step $k$, where $avail_{m}=1$ denotes that vehicle $v_m$ is available for offering its computation resource or processing current subtask, and $avail_{m}=0$ otherwise. Also, $\mathcal{O}^{(k)}= \{(x_m,y_m)\mid v_m \in \mathcal{V}\}$ is the instantaneous location of vehicles at decision step $k$.

\subsection{Action Space}
During each decision step $k$, we need to determine which vehicle should be assigned to each subtask based on the system state $s^{(k)}$ and subtask scheduling priority list $\mathcal{L}^{\mathsf{rank}}$. In particular, at decision step $k$, for current subtask $b_{\tau(k)}$, action $a^{(k)}$ is defined as
\vspace{-0.5em}
\begin{flalign}	
	a^{(k)} \in \{1,2,\cdots,|\mathcal{V}|\}, \label {eq_26}
\end{flalign}
where $a^{(k)}=1$ implies that current subtask $b_{\tau(k)}$ is processed locally on task owner $v_1$, and $a^{(k)} \in \{2,\cdots,|\mathcal{V}| \}$ implies that current subtask $b_{\tau(k)}$ is allocated to other vehicles for a faster execution.
\vspace{-0.3em}
\subsection{Reward Design}
At decision step $k$, given state $s^{(k)}$, we associate performing action $a^{(k)}$ for allocating of current subtask $b_{\tau(k)}$ to an immediate reward $r^{(k)}$ leveraged to evaluate the quality of action $a^{(k)}$. We define the reward $r^{(k)}$ as the decrease in the $EFT$ of all subtasks as
\vspace{-0.3em}
\begin{flalign}
	\hspace{-3em}  r^{(k)} =\underbrace{\max\limits_{b_i \in \mathcal{B}, v_m \in \mathcal{V}}\{EFT^{(k-1)}_{i,m} \}}_{(\text{I})}-\underbrace{\max\limits_{b_i \in \mathcal{B}, v_m \in \mathcal{V}}\{EFT^{(k)}_{i,m} \}}_{(\text{II})}, \hspace{-2em}\label {eq_27} 
\end{flalign}
where term (I) and (II) denote the maximum DAG task completion time before and after scheduling the current subtask, respectively. We next demonstrate the rationality of reward function introduced above. 
\begin{figure}[!t]
	\centering
	\includegraphics[width=3.5in]{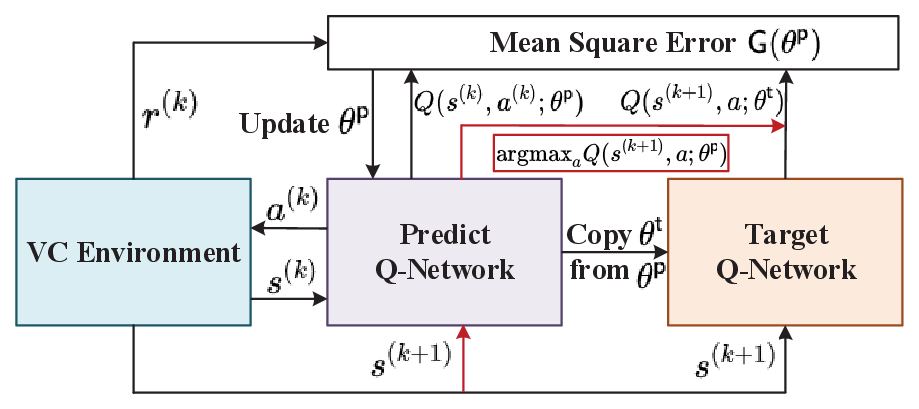}
	\caption{Flow of Double DQN methodology.}
	\label{fig_6}
	\vspace{-0.5cm}
\end{figure}

\textbf{Rational of the Choice of  Reward.} Let $K$ denote the total number of decision steps. According to (\ref{eq_27}), the discounted cumulative reward can be calculated as
\begin{flalign}
	&R = \sum_{k=1}^{K} \gamma_1^{k}r^{(k)} \nonumber \\
	&\hspace{-0.5em}=\hspace{-0.5em} \sum_{k=1}^{K} \gamma_1^{k}\Big( \max _{b_i \in \mathcal{B}, v_m \in \mathcal{V}}\{{EFT}^{(k-1)}_{i,m} \}-\hspace{-1em}\max _{b_i \in \mathcal{B}, v_m \in \mathcal{V}}\{{EFT}^{(k)}_{i,m} \}\hspace{-0.2em}\Big), \hspace{-0.5em}\label {eq_28} 
\end{flalign}
where $\gamma_1$ is the discount factor. Assuming $\gamma_1=1$ for simplicity, since at decision step $k$, we determine the allocation of only the current subtask $b_{\tau(k)}$ according to scheduling priority list $\mathcal{L}^{\mathsf{rank}}$, we have $K=|\mathcal{L}^{R}|=|\mathcal{B}|$. Thus, (\ref{eq_28}) can be rewritten as
 \begin{flalign}
 	&R = \sum_{k=1}^{K}\Big(\max _{v_m \in \mathcal{V}}\{{EFT}_{\tau(k-1),m} \}-\max _{v_m \in \mathcal{V}}\{{ EFT}_{\tau(k),m} \}\Big) \nonumber \\
 	&=\Big(\max _{v_m \in \mathcal{V}}\{{EFT}_{\tau(0),m} \}-\max _{v_m \in \mathcal{V}}\{{EFT}_{\tau(1),m} \} \nonumber\\
 	&+\max _{v_m \in \mathcal{V}}\{{EFT}_{\tau(1),m} \}+\cdots -\max _{v_m \in \mathcal{V}}\{{EFT}_{\tau(K),m} \} \Big) \nonumber \\
 	&=-\Big(\max _{v_m \in \mathcal{V}}\{{EFT}_{\tau(K),m} \}-\max _{v_m \in \mathcal{V}}\{{EFT}_{\tau(0),m} \}\Big),  \label {eq_29} 
 \end{flalign}
where we define $b_{\tau(0)}$ as the virtual subtask with $\max _{v_m \in \mathcal{V}}\{{EFT}_{\tau(0),m}\}=0$. The last result in (\ref{eq_29}) (i.e., term $-\max _{v_m \in \mathcal{V}}\{{EFT}_{\tau(K),m} \}$) indicates that maximizing the cumulative reward is consistent with minimizing the task completion time given in (\ref{eq_9}).

Hereafter, in order to solve the above mentioned MDP, we resort to a DDQN, which adopts the action (i.e., subtask-to-vehicle allocation) at each decision step yielding the largest Q-value (i.e., state-action value) prior to DAG task scheduling over dynamic VCs.

\subsection{Double Deep Q-Network}
\subsubsection{Deep Q-network} We first describe DQN methodology\cite{b49}, which paves the way for DDQN. In DQN, we have two deep neural networks (DNNs) called \emph{predict} Q-network $Q(s,a; \theta^{\mathsf{p}})$ and \emph{target} Q-network $Q(s,a; \theta^{\mathsf{t}})$. Particularly, $\theta^{\mathsf{p}}$ and $\theta^{\mathsf{t}}$ are the vectors of weights/parameters of DNNs, and $s$ and $a$ denote the state and action, respectively.

\textbf{Predict Q-value.} At each decision step $k$, given state $s^{(k)}$, using the predict Q-network, the DRL agent first estimates/predicts the Q-value $Q(s^{(k)}, a;\theta^{\mathsf{p}})$ of all actions $a = 1,2,\cdots,|\mathcal{V}|$, where $s^{(k)}$ consists of the extracted feature of current subtask $b_{\tau(k)}$ and vehicles' parameters given in (\ref{eq_25}). Q-value is a measure of the quality of the action: a higher Q-value is an indicator to a better action.

\textbf{Action selection.} The DRL agent then performs an action $a^{(k)}$ using a max mathematical estimator as follow 
\begin{flalign}
a^{(k)}=\text{argmax}_{a}{Q}(s^{(k)}, a; \theta^{\mathsf{p}}), a\in \{1,2\cdots|\mathcal{V}|\}.\label {eq_36} 
\end{flalign}
The DRL agent then receives a reward $r^{(k)}$ computed by (\ref{eq_27}). 

\textbf{Target Q-value.} The system subsequently transits to the next state $s^{(k+1)}$, and the DRL agent resorts to target Q-network for calculating the target Q-value of state $s^{(k)}$, denoted by $\mathsf{y}^{(k)}$:
\begin{flalign}
	&\hspace{-0.5em}\mathsf{y}^{(k)}\!=\! r^{(k)} + \gamma_2 \underbrace{Q\left(s^{(k+1)}, \underbrace{\text{argmax}_{a}Q\left(s^{(k+1)}, a;\theta^{\mathsf{t}}\right)}_{(\text{I})}; \theta^{\mathsf{t}}\right)}_{(\text{II})}, \nonumber\\
	&\quad\quad\quad\quad\quad\quad\quad\quad\quad \quad\quad\quad\quad\quad\quad a\in \{1,2\cdots|\mathcal{V}|\}.\hspace{-0.5em}\label {eq_30} 
\end{flalign}

To obtain the parameter $\theta^{\mathsf{p}}$, the mean square error, denoted by $\mathsf{G}(\theta^{\mathsf{p}})$, is used with discount factor $\gamma_2$ as follows
\begin{flalign}
	\mathsf{G}(\theta^{\mathsf{p}}) \!=\!\! \frac{1}{2}\left[\mathsf{y}^{(k)}- Q(s^{(k)},a^{(k)}; \theta^{\mathsf{p}})\right]^{2}\!\!\!, a\in \{1,2\cdots|\mathcal{V}|\}.\label {eq_31} 
\end{flalign}
Also, the weights of the target network $\theta^{\mathsf{t}}$ are periodically copied from the predict network $\theta^{\mathsf{p}}$.

\subsubsection{Double Deep Q-network} In standard DQN, the $\max$ operator employs the same values to both select (i.e., term (I) in (\ref{eq_30})) and evaluate (i.e., term (II) in (\ref{eq_30})) an action. This implies that the Q-values are updated based on estimated future rewards, rather than actual rewards. Thus, there is a risk of overestimating Q-values, especially when the estimates are based on an inaccurate model of the environment. To prevent this, we resort to DDQN \cite{b50} aiming at separating the action selection from the action evaluation. In DDQN, the action with the maximum Q-value is selected using the predict network, and the Q-value for this action is evaluated using the target network. In particular, DDQN uses the same approach for predicting Q-value and selecting action as DQN. However, it uses a different target network update rule detailed next.

\textbf{Target Q-value in DDQN.} The target value of state $s^{(k)}$ in DDQN (see the red line shown in Fig. \ref{fig_6}), denoted by $\mathsf{y}^{(k)}_{\mathsf{Double}}$ is changed from (\ref{eq_30}) to 
\begin{flalign}
	\hspace{-0.5em} &\mathsf{y}^{(k)}_{\mathsf{Double}}\!=\! r^{(k)} \!+\! \gamma_2{Q}\left(\!s^{(k+1)}, \underset{a}{\arg \max } Q\left(s^{(k+1)}, a, \theta^{\mathsf{p}}\!\right)\!;  \theta^{\mathsf{t}}\right), \nonumber \\ 
	&\quad\quad\quad\quad\quad\quad\quad\quad\quad\quad\quad\quad\quad\quad a\in \{1,2\cdots|\mathcal{V}|\},\hspace{-0.5em}\label {eq_32} 
\end{flalign}
where the action $a$ is conducted by predict Q-network $Q(s, a;\theta^{\mathsf{p}})$. Finally, the mean square error for training predict Q-network $Q(s, a;\theta^{\mathsf{p}})$ is modified from (\ref{eq_31}) to  
\begin{flalign}
	\hspace{-1em}\mathsf{G}(\theta^{\mathsf{p}}) \!=\!\!\frac{1}{2}\Bigg[\!\mathsf{y}^{(k)}_{\mathsf{Double}} - Q(s^{(k)},a^{(k)}; \theta^{\mathsf{p}})\!\Bigg]^{2}\!\!, a\in \{1,2\cdots|\mathcal{V}|\}.\hspace{-1em} \label {eq_33} 
\end{flalign}
Using which the parameter $\theta^{\mathsf{p}}$ is obtained, the weights of the target network $\theta^{\mathsf{t}}$ are then periodically copied from the predict network $\theta^{\mathsf{p}}$.
\begin{algorithm} [!t]
	\small
	\centering 
	\caption {Training Process for GA-DRL}\label{algorithm_2}
	\begin{algorithmic}[1]
		\STATE \textbf{Input:} GAT structure, DDQN structure, learning rate $\mu$, and decision step $\mathcal{K}$ for copying $\theta^{\mathsf{t}}$ from $\theta^{\mathsf{p}}$
		\STATE \textbf{Output:} GNN with parameters $\mathcal{W}$ and $\mathcal{A}$, DDQN with parameters $\theta^{\mathsf{p}}$ and $\theta^{\mathsf{t}}$
		\FOR {each training episode}
		\STATE Consider a DAG task $\mathcal{G}=(\mathcal{B},\mathcal{E})$ and a VC $\mathcal{V}$
		\STATE Obtain list $\mathcal{L}^{rank}$ using (\ref{eq_18})- (\ref{eq_22})
		\FOR {decision step $k = 1 \cdots K$}
		\STATE For current subtask $b_{\tau(k)}$ use Algorithm \ref{algorithm_1} to extract its feature $h^{(L)}_{\tau(k)}$. 
		\STATE The predict Q-network takes $s^{(k)}=\Big\{h^{(L)}_{\tau{(k)}},\mathcal{I}^{(k-1)}, \mathcal{A}^{(k)}, \mathcal{O}^{(k)}\Big\}$ and performs an action $a^{(k)}$ according to $\epsilon$-greedy policy
		\STATE Next decision step $s^{(k+1)}$ is obtained, and DRL agent receives a reward $r^{(k)}$ in (\ref{eq_27})
		\STATE Store $\{s^{(k)}, a^{(k)}, r^{(k)}, s^{(k+1)}\}$ in the buffer $\mathcal{R}$
		\STATE Sample mini-batch $\mathcal{D}$ from $\mathcal{R}$ uniformly at random
		\STATE Use $\mathcal{D}$ to train GAT with parameters $\mathcal{W}$, $\mathcal{A}$ and DDQN with $\theta^{\mathsf{p}}$ and $\theta^{\mathsf{t}}$ jointly by minimizing (\ref{eq_35})
		\STATE Update GAT and predict Q-network every step
		\STATE Set target Q-network to be a copy of the predict Q-network every $\mathcal{K}$ decision steps
		\ENDFOR
		\ENDFOR	
	\end{algorithmic}	
\end{algorithm}

\subsection{Training Process}
We consider training of DRL through a series of \emph{episodes}, where each episode contains total of $K$ sequential decision steps. At each decision step $k$, DRL agent generates a pair of observation $(s^{(k)},a^{(k)},r^{(k)},s^{(k+1)})$. An episode is considered to be complete when a vehicle is assigned the subtask with the lowest scheduling priority, which is listed in the last position of $\mathcal{L}^{\mathsf{rank}} $ (i.e, $K=|\mathcal{B}|$).

\subsubsection{Q-network training} 
Based on policy gradient algorithm\cite{b40}, predict Q-network $Q(s, a;\theta^{\mathsf{p}})$ is trained by iteratively tuning the weights $\theta^{\mathsf{p}}$ at each decision step $k$ through minimizing the mean square error given in (\ref{eq_33}) as follows:
\begin{flalign}
	\theta^{\mathsf{p}} \leftarrow \theta^{\mathsf{p}}- \mu\frac{\partial\mathsf{G}(\theta^\mathsf{p})}{\partial\theta^{\mathsf{p}}} \label {eq_34} 
\end{flalign}
where $\mu$ is the tunable learning rate. As for the target Q-network $Q(s, a;\theta^{\mathsf{t}})$, $\theta^{\mathsf{t}}$ is copied from $\theta^{\mathsf{p}}$ at beginning, and $\theta^{\mathsf{t}}$ will be iteratively updated to $\theta^{\mathsf{p}}$ after conducting some iterations (5 decision steps in our simulations).

We adopt a $\epsilon$-greedy policy to select action, in which the DRL agent probabilistically explores the actions which have not been adopted yet instead of an action with the maximum Q-value in (\ref{eq_36}). Also, we leverage a replay buffer $\mathcal{R}$ to store the sequence of $(s^{(k)},a^{(k)},r^{(k)},s^{(k+1)})$ obtained through decision steps $k$. In particular, the gradient in (\ref{eq_34}) is obtained by selecting mini-batches of data from the reply buffer. At each decision step $k$, we consider the feasibility of actions for the current subtask $b_{\tau(k)}$. Actions that meet constraint (C5) are defined as feasible, while the others are infeasible. We leverage action mask\cite{b47} technique to prevent DDQN from performing infeasible actions. In this approach, the Q-value for an infeasible action is set to a large negative value, to ensure that taken actions are feasible.

\subsubsection{GAT training} The state $s^{(k)}$ which consists of the extracted features of current subtask $b_{\tau(k)}$ is obtained from the GAT with parameters $\mathcal{W}=\{W^{(\ell)(z)} \mid 1\leq \ell \leq L,\ 1\leq z\leq Z\}$ and $\mathcal{A}=\{A^{(\ell)(z)} \mid 1\leq \ell \leq L,\ 1\leq z\leq Z\}$. Thus, we can rewrite the right hand-side of (\ref{eq_33}) as
\begin{flalign}
	&\!\!\Bigg[r^{(k)} \!+\! \gamma_2{Q}\left(\!s^{(k+1)}(\mathcal{W},\mathcal{A}), \underset{a^{*}}{\arg \max } Q\left(s^{(k+1)}, a^{*}, \theta^{\mathsf{p}}\!\right)\!;  \theta^{\mathsf{t}}\right)\nonumber\\
	&\quad\quad - Q(s^{(k)}(\mathcal{W},\mathcal{A}),a^{(k)}; \theta^{\mathsf{p}})\Bigg]^{2},  a\in \{1,2\cdots|\mathcal{V}|\},\label {eq_35} 
\end{flalign}
which indicates that parameters $\mathcal{W}$, $\mathcal{A}$ and $\theta^{\mathsf{p}}$ are trained simultaneously by minimizing (\ref{eq_35}) during the decision steps of the DRL agent.

\textbf{Algorithm} \ref{algorithm_2} presents a pseudocode of GA-DRL training procedure.

\section{Performance Evaluation}
In this section, we first provide parameter settings for simulations. We then study the convergence of GA-DRL. Finally, we compare the performance of GA-DRL with four DAG task scheduling benchmarks in terms of the task completion time.

\subsection{Simulation Setting}
\begin{figure}[bthp]
	\centering
	\subfigure[Real-world traffic region.]{
		\begin{minipage}[t]{0.5\linewidth}
			\centering
			\includegraphics[width=\linewidth]{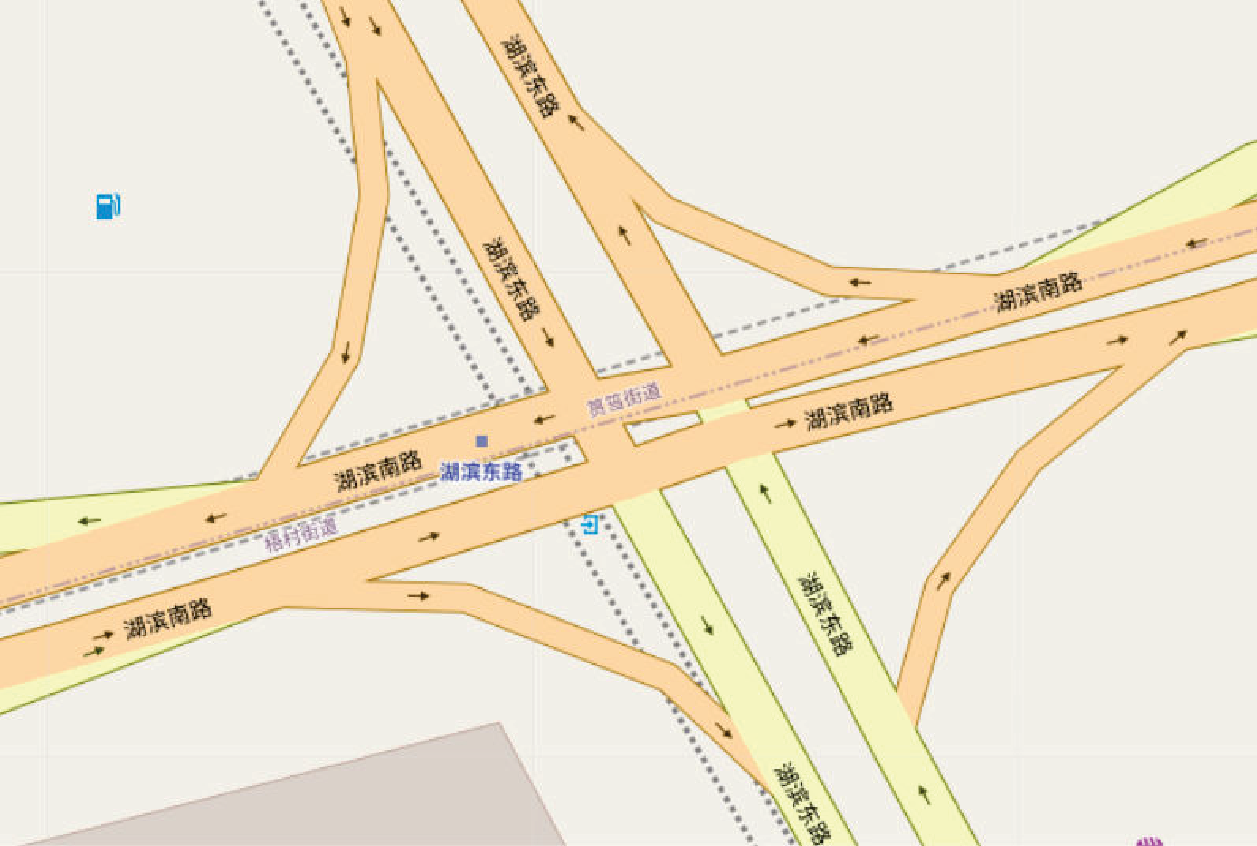}
		\end{minipage}%
	}%
	\subfigure[Vehicle trajectory.]{
		\begin{minipage}[t]{0.5\linewidth}
			\centering
			\includegraphics[width=0.85\linewidth]{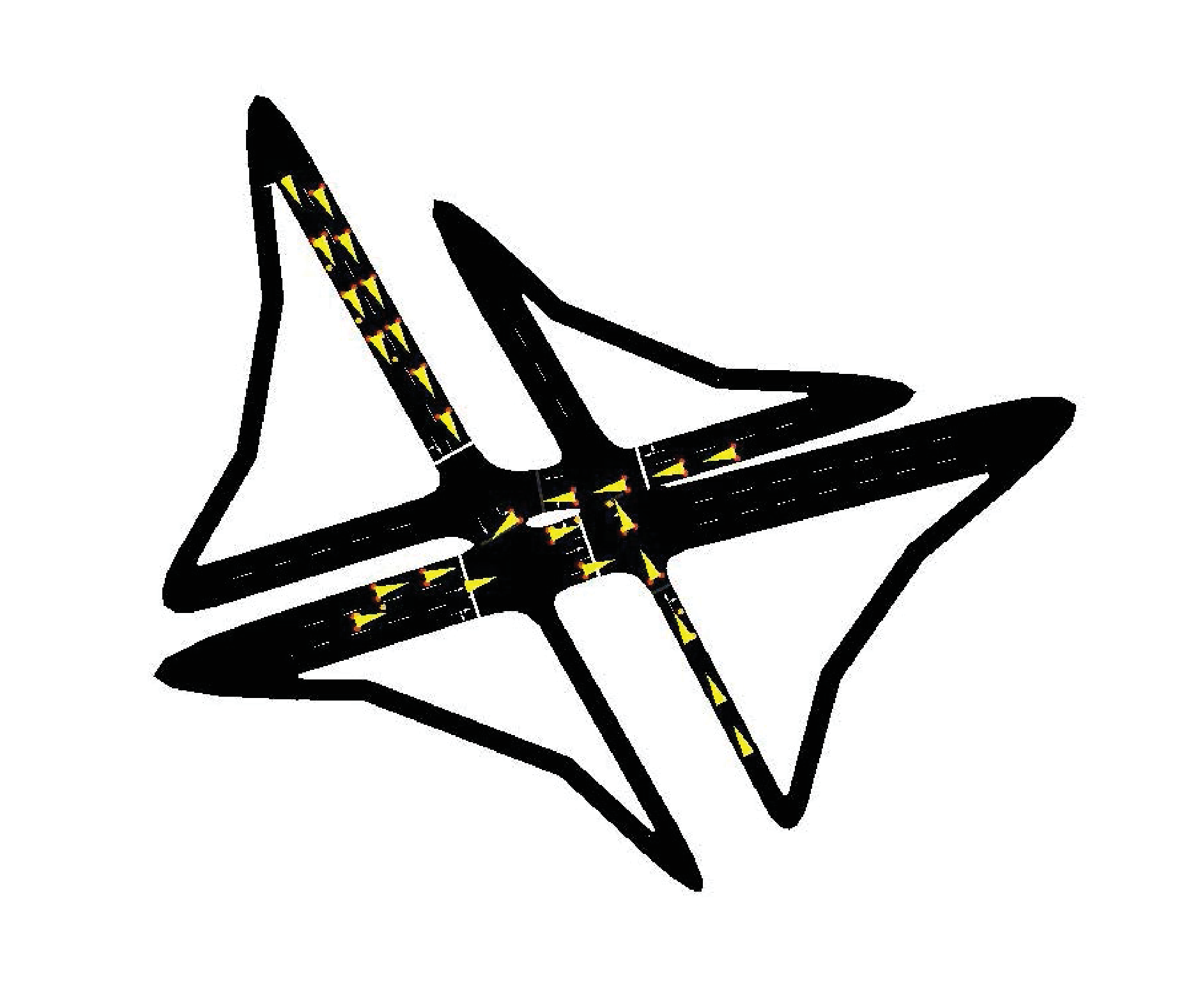}
		\end{minipage}%
	}%
	\centering
	\caption{VC network visualization.}
	\label{fig_7}
	\vspace{-0.2cm}
\end{figure}

\textbf{Simulation environment.} All neural networks considered in this work are implemented using PyTorch 2.0.0\cite{b44} and Python 3.8.1 platforms, and Adam\cite{b45} is leveraged to optimize networks. In our simulations, we consider a real-world highway traffic region as shown in Fig. \ref{fig_7}(a) of size 1km$\times$1km in Xiamen, China, obtained from OpenStreetMap\cite{b10}. Moreover, SUMO\cite{b11} is utilized to import mobile vehicles using the mobility model developed in (\ref{eq_1})-(\ref{eq_2}), and subsequently emulate a real-world VC as shown in Fig. \ref{fig_7}(b). Also, the arrival time of each vehicle, i.e., $AT_m$, is assumed to be uniformly distributed in $[1,5]$ (in second) for analytical simplicity, and $\mu_g=50$ (in Kilometres per hour) with $\sigma_g=10$.

\textbf{Parameter setting of DAG tasks.} The task owner has a DAG task which is generated according to \cite{b5}. We assume that the computation capability of each vehicle is uniformly distributed in $[1, 10]$ (in GHz)\cite{b19}, the distance between different vehicles during the task scheduling process are captured by SUMO, and function $\Psi(\cdot)$ in (\ref{eq_6}) is defined as $\Psi\left(PL\left(d_{m, n}(t)\right)\right) = 0.15 PL\left(d_{m, n}(t)\right) + 0.001$\cite{b38}. Also, the computation workload of each subtask is uniformly distributed in $[1, 2]$ (in Gigaclock cycles)\cite{b26} and the transmission data size of each edge is uniformly distributed in $[100, 500]$ (in KB)\cite{b26}. During training, we have chosen $\epsilon$-greedy policy with $\epsilon=0.9$ and discount factor $\gamma_2 = 0.9$.

\subsection{Convergence Performance}
\begin{figure}[bthp]
	\centering
	\includegraphics[width=3.5in]{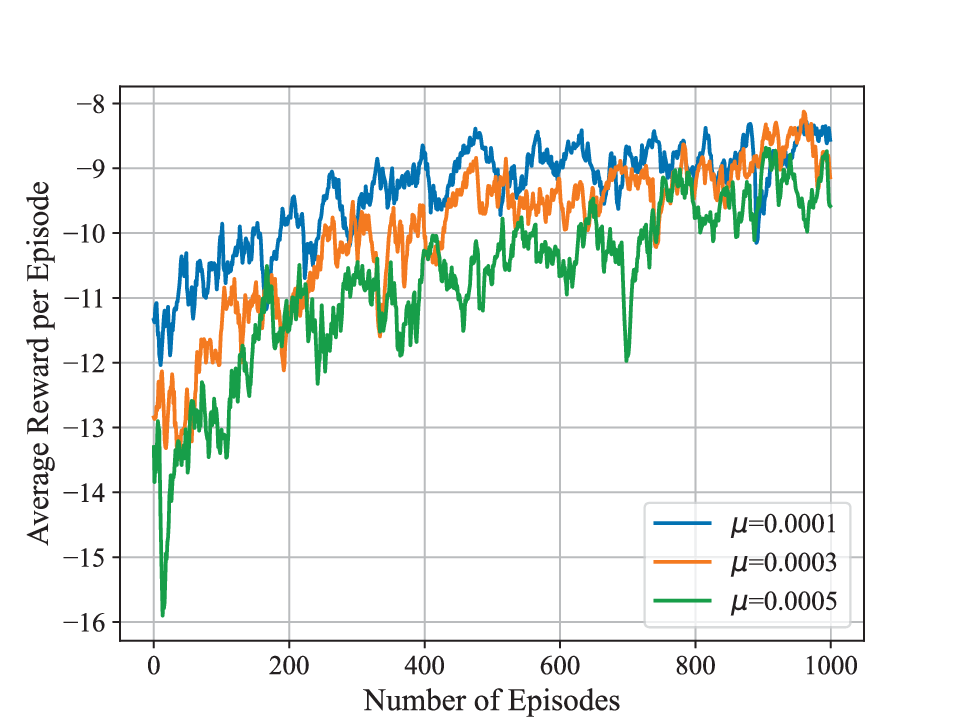}
	\caption{Average reward under different learning rates.}
	\label{fig_8}
\end{figure}
In Fig. \ref{fig_8}, we depict the convergence behavior of GA-DRL with respect to the number of episodes. Note that the best convergence and reward values are achieved when the GA-DRL's learning rate is 0.0001. On the other hand, as the learning rate increases from 0.0001 to 0.0005, the average reward is significantly decreased due to the instability of learning. As a result, we fix the learning rate of the GA-DRL to 0.0001 when comparing it with  benchmarks in the following.

\subsection{Benchmarks}
To study the performance of GA-DRL, we implement four DAG task scheduling benchmarks, including LPS, HEFT \cite{b12}, MGA \cite{b15}, and DRLOSM \cite{b46} as detailed below.

\begin{itemize}
	\item \emph{Local processing scheme (LPS):} All subtasks are processed locally by the task owner itself without offloading to other vehicles.
	\item \emph{Heterogeneous earliest finish time (HEFT)\cite{b12}:} All subtasks are first sorted according to their ranking value in (\ref{eq_24}). The subtasks are assigned to the vehicles that can complete them in the shortest time. The HEFT algorithm does not take into account the constraint of V2V transmission (C5) since it was designed for a static computing environment.  We assume that subtasks-to-vehicles allocations that do not satisfy constraint (C5) are executed locally.
	\item \emph{Modified genetic algorithm (MGA)\cite{b15}:} MGA considers an integer encoding to denote subtask-to-vehicle assignments. The assignments with high fitness (i.e., low task completion time) are stochastically selected to perform crossover (i.e., exchange their processing vehicles). Finally, a mutation (i.e., changing the processing vehicle) is adopted to avoid early convergence. MGA considers a VC environment satisfying  V2V communication constraint (C5). 
	\item  \emph{DRL offloading scheduling method (DRLOSM)\cite{b46}:} DRLOSM is an improved version of the method proposed in \cite{b46}. All subtasks are first sorted according to their ranking value in (\ref{eq_24}). DRLOSM uses a DDQN architecture, where at decision step $k$, the raw feature of \emph{current subtask} $b_{\tau(k)}$ is integrated in $s^{(k)}$ without the use of GNNs. DRLOSM also satisfies the V2V communication constraint (C5) through an action mask module. 
\end{itemize}

\subsection{Simulation Results of Randomly Generated DAG Tasks}
We conduct performance evaluations by analyzing the average completion time of DAG tasks for various numbers of layers\footnote{The number of layers of a DAG task refers to the length of the longest path from the starting subtask to the finishing subtask. For a DAG task with a fixed number of subtasks, as the number of layers increases/decreases, there are more/less subtasks that are successors of the same subtask, implying a higher/lower potential for parallelism during the task execution.} of DAG tasks, subtasks, and vehicles in the network. The results are the average performance obtained via 100 independent Monte-Carlo iterations. Also, to compare the generalizability of DRLOSM and our GA-DRL, during the training period, we use the same DAG task topology, while deploying them for various DAG task topologies under performance evaluation.

\subsubsection{Impact of the number of vehicles in VC} 
\begin{figure}[!t]
	\centering
	\includegraphics[width=3.5in]{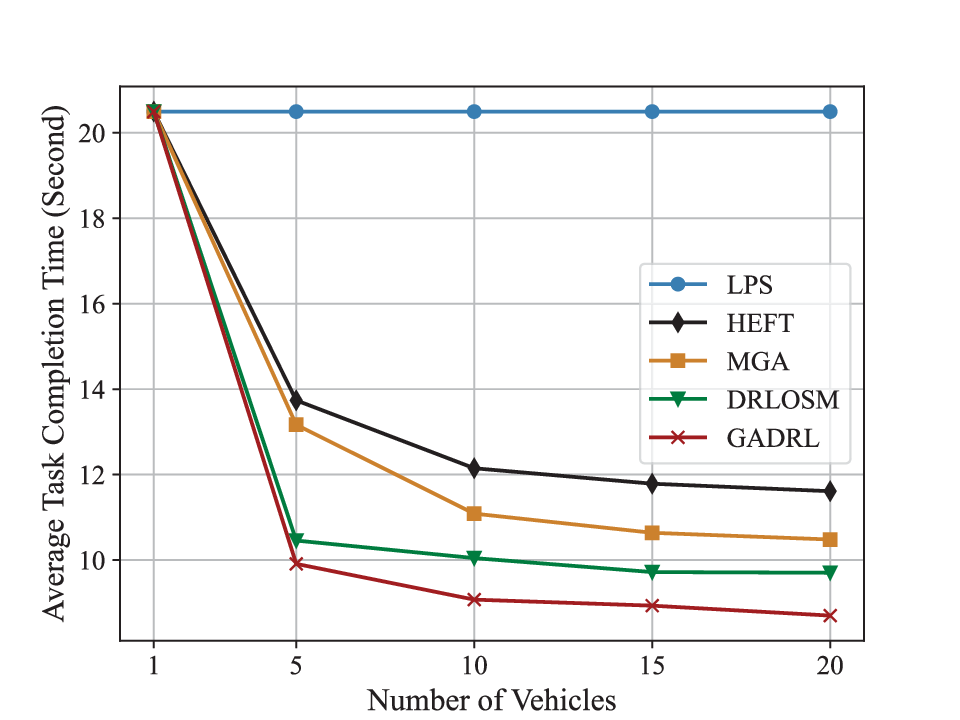}
	\caption{Performance evaluations upon considering the different numbers of vehicles within VC.}
	\label{fig_9}
	\vspace{-0.5cm}
\end{figure}

The results presented in Fig. \ref{fig_9} illustrate the impact of increasing the number of vehicles from 1 to 20 on the completion time of the DAG task. The experiment was conducted with 20 subtasks and 5 layers. We observed that when only one vehicle is involved in VC, all DAG subtasks have to be executed sequentially and locally, resulting in the same completion time across different algorithms. However, as the number of vehicles increases, the completion of DAG tasks is significantly accelerated due to the sufficient computation resources. Overall, GA-DRL outperforms other algorithms in terms of task completion time. Itis 51.63$\%$ better than LPS, 27.82$\%$ better than HEFT, 24.69$\%$ better than MGA, 5.17$\%$ better than DRLOSM at 5 vehicles; and is 57.59$\%$ better than LPS, 25.15$\%$ better than HEFT, 17.08$\%$ better than MGA, and 10.41$\%$ better than DRLOSM at 20 vehicles.
\begin{figure}[!t]
	\centering
	\includegraphics[width=3.5in]{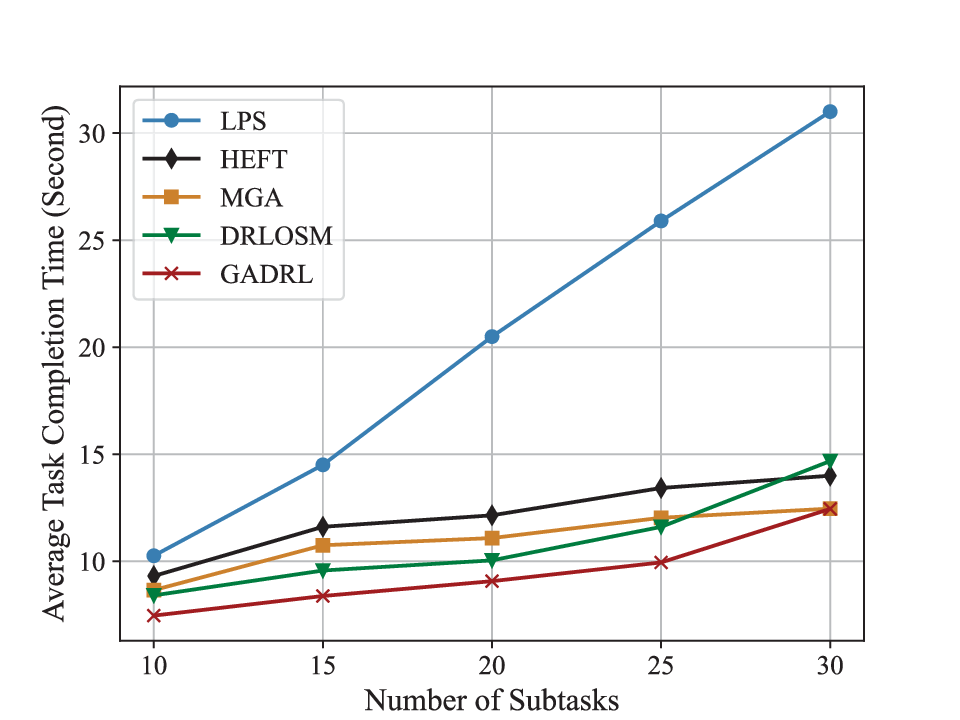}
	\caption{Performance evaluations upon considering the different numbers of subtasks within DAG task.}
	\label{fig_10}
	\vspace{-0.5cm}
\end{figure}
\begin{figure}[!t]
	\centering
	\includegraphics[width=3.5in]{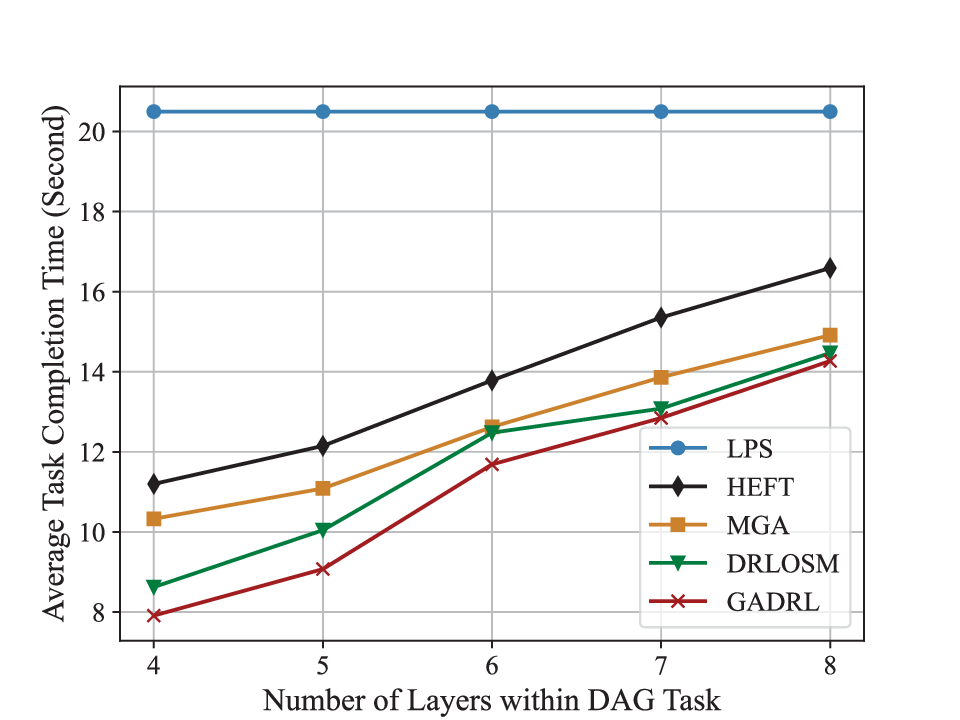}
	\caption{Performance evaluations upon considering the different numbers of layers within DAG task.}
	\label{fig_11}
	\vspace{-0.5cm}
\end{figure}
\subsubsection{Impact of the number of subtasks} In Fig. \ref{fig_10}, we can see the evaluation of the completion time for DAG tasks as the number of subtasks increases. In this experiment, we set the number of vehicles involved in the VC at 10, and the number of layers at 5. The results show that our proposed GA-DRL algorithm outperforms the other four benchmarks, achieving faster task completion times. Additionally, Fig. \ref{fig_10} demonstrates the effectiveness of GA-DRL compared to conventional DRL in terms of generalizability. As the number of subtasks increases from 25 to 30, the task completion time of DRLOSM becomes longer than that of both MGA and HEFT. This is due to the fact that the topologies of DAG tasks become more complicated, making the DRLOSM algorithm, which relies solely on human-selected features without the usage of GNNs, unable to capture the topological information of the newly generated DAG task topologies. On the other hand, our GA-DRL algorithm benefits from the subtasks’ features, which are automatically learned from GAT, making its models well generalizable to unseen DAG task topologies. In summary, the performance of GA-DRL in terms of the task completion time is 27.29$\%$ better than LPS, 19.87$\%$ better than HEFT, 13.76$\%$ better than MGA, 11.29$\%$ better than DRLOSM at 10 subtasks; and is 59.84$\%$ better than LPS, 11.01$\%$ better than HEFT, 0.08$\%$ better than MGA, and 15.19$\%$ better than DRLOSM at 30 subtasks.

\subsubsection{Impact of the number of layers within DAG task} In Fig. \ref{fig_11}, it is evident that changing the number of DAG task layers from 4 to 8 has a significant impact on the completion time of the DAG task. In this result, we considered 20 subtasks and 10 vehicles. It is observed that increasing the number of layers leads to a longer completion time. This is because, as the number of layers increases, the parallelism of the DAG task decreases, resulting in more subtasks being executed in a sequential manner. This, in turn, leads to a longer task completion time. The performance of GA-DRL in terms of the task completion time is 61.39$\%$ better than LPS, 29.31$\%$ better than HEFT, 23.35$\%$ better than MGA, 8.27$\%$ better than DRLOSM at 4 layers; and is 30.41$\%$ better than LPS, 14.04$\%$ better than HEFT, 4.36$\%$ better than MGA, and 1.38$\%$ better than DRLOSM at 8 layers.

\subsection{Simulation Results for Real Application DAG Task}
\begin{figure}[!t]
	\centering
	\includegraphics[width=3.5in]{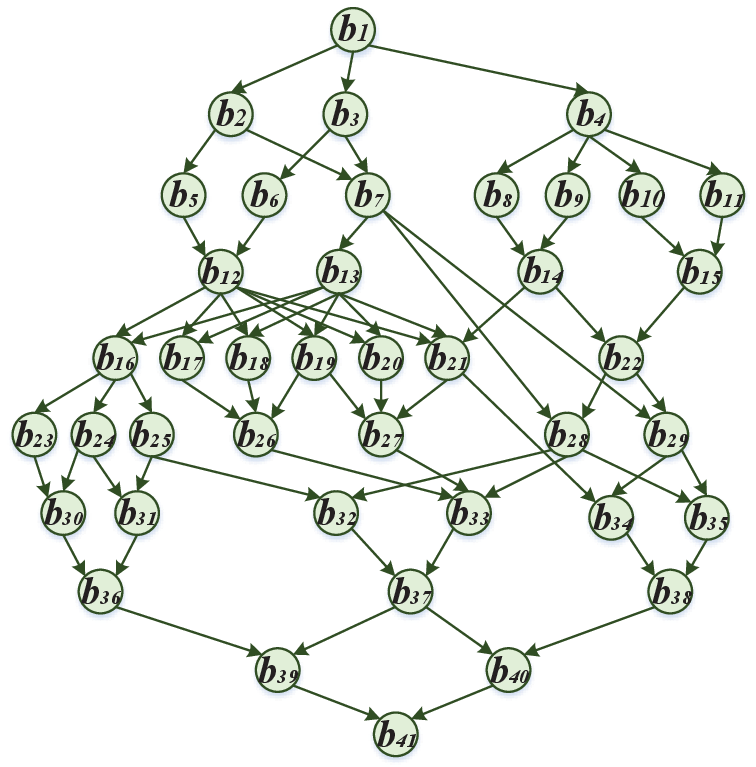}
	\caption{The DAG task model of the molecular dynamics code\cite{b12}.}
	\label{fig_12}
\end{figure}

In Fig. \ref{fig_12}, we illustrate a real-world DAG task of a modified molecular dynamic code \cite{b12}. The subtasks' computation workload and transmission data size were set according to the parameter settings, and we considered 20 vehicles in the result. Table \ref{table3} presents the performance comparison of various benchmarks, except LPS\footnote{LPS is excluded, since it takes no algorithm running time, while we have demonstrated above that the task completion time of LPS is always worse than that of other benchmarks.}, with respect to the DAG task completion time (in seconds) and the algorithm running time (in seconds). It is important to note that DRLOSM, which solely learns from human-selected features of subtasks without the usage of GAT, exhibits a higher task completion time than others, such as HEFT, MGA, and our GA-DRL. This is a clear indication of the superiority of generalization of our GA-DRL, especially in the case of a large number of subtasks. Additionally, MGA has the longest running time among the benchmarks due to its internal iteration time for convergence. However, our GA-DRL shows better performance in terms of task completion time at the mild cost of higher algorithm running time. The performance of GA-DRL is $19.09\%$ better than HEFT, $4.31\%$ better than MGA, and $25.64\%$ better than DRLOSM. In summary, simulation results verify that our proposed GA-DRL algorithm offers an efficient and commendable reference in scheduling DAG tasks over dynamic VCs.

\section{Conclusion}
In this paper, we focused on scheduling DAG tasks using a combination of GNNs and DRL. We approached the problem by modeling it as an MDP, and using a GAT module to extract features for each subtask in the DAG task topology. We then integrated the GAT with a DDQN to allocate subtasks to vehicles while taking into account the dynamics and heterogeneity of the vehicles. Our GAT uses multiple heads to enhance information for important subtasks and aggregates topological information from both subtasks' predecessors and successors. We also incorporated a non-uniform neighborhood sampling methodology to improve the GAT's generalizability. Our evaluations showed that our GA-DRL method outperforms benchmarks in terms of task completion time. Future work could explore cooperation among different vehicles and optimizing start times for network flows to transmit the data of subtasks among each other.

\begin{table}[!t]
	\renewcommand{\arraystretch}{1.3}
	\setlength\tabcolsep{5pt}
	\caption{Performance Comparison Under Different Metrics For Molecular Dynamics Code DAG}
	\label{table3}
	\centering
	\begin{tabular}{|c|c|c|c|c|}
		\hline
		\textbf{Metric}                 & \textbf{HEFT} & \textbf{MGA} & \textbf{DRLOSM} & \textbf{GA-DRL} \\ \hline
		Task completion time          & 21.474      & 18.157       & 23.363            & 17.373        \\ \hline
		Algorithm running time  & 2.4542        & 8.2229       & 2.0755             & 2.9119        \\ \hline
	\end{tabular}
	\vspace{-0.4cm}
\end{table}


\vspace{-0.5em}
\begin{thebibliography}{00}
\bibitem{b1} Z. Chen \emph{et al.}, “An empirical study of latency in an emerging class of edge computing applications for wearable cognitive assistance,” in \emph{Proc. ACM/IEEE Symp. Edge Comput.}, Oct. 2017, pp. 1–14.
\bibitem{b2} T. Taleb, A. Ksentini, M. Chen, and R. Jantti, “Coping With Emerging Mobile Social Media Applications through Dynamic Service Function Chaining,” \emph{IEEE Trans. Wireless Commun.}, vol. 15, no. 4, pp. 2859–2871, Apr. 2016. 
\bibitem{b3} L. F. Bittencourt, R. Sakellariou and E. R. M. Madeira, “DAG scheduling using a lookahead variant of the heterogeneous earliest finish time algorithm,” in \emph{Proc. Eur. Conf. Parallel Process.}, Apr. 2010, pp. 27-34.
\bibitem{b4} G. C. Sih and E. A. Lee, “A Compile-Time Scheduling Heuristic for Interconnection-Constrained Heterogeneous Processor Architectures,” \emph{IEEE Trans. Parallel Distrib. Syst.}, vol. 4, no. 2, pp. 175-187, Feb. 1993.
\bibitem{b5} H. Arabnejad and J. G. Barbosa, “List Scheduling Algorithm for Heterogeneous Systems by An Optimistic Cost Table,” \emph{IEEE Trans. Parallel Distrib. Syst.}, vol. 25, no. 3, pp. 682-694, Mar. 2014.
\bibitem{b6} H. Kanemitsu, M. Hanada and H. Nakazato, “Clustering-Based Task Scheduling in A Large Number of Heterogeneous Processors,” \emph{IEEE Trans. Parallel Distrib. Syst.}, vol. 27, no. 11, pp. 3144-3157, Nov. 2016.
\bibitem{b7} Z. Liu, M. Liwang, S. Hosseinalipour, H. Dai, Z. Gao and L. Huang, “RFID: Towards low latency and reliable DAG task scheduling over dynamic vehicular clouds," \emph{IEEE Trans. Veh. Technol.}, Early Access, Apr. 2023.
\bibitem{b8} M. Barbera, S. Kosta, A. Mei, and J. Stefa, “To Offload or Not To Offload? The Bandwidth and Energy Costs of Mobile Cloud Computing,” in \emph{Proc. IEEE Conf. Comput. Commun.}, Apr. 2013, pp.
1285–1293.
\bibitem{b9} W. Shi, J. Cao, Q. Zhang, Y. Li and L. Xu, “Edge Computing: Vision
and Challenges,” \emph{IEEE Internet Things J.}, vol. 3, no. 5, pp. 637-646,Oct. 2016.
\bibitem{b10} M. Haklay and P. Weber, “OpenStreetMap: User-Generated Street Maps,” \emph{IEEE Pervasive Comput.}, vol. 7, no. 4, pp. 12-18, Oct. 2008.
\bibitem{b11} P. A. Lopez et al., “Microscopic traffic simulation using SUMO,” in \emph{Proc. Int. Conf. Intell. Transp. Syst. (ITSC)}, Nov. 2018, pp. 2575-2582.
\bibitem{b12} H. Topcuoglu, S. Hariri and Min-You Wu, “Performance-Effective and Low-Complexity Task Scheduling for Heterogeneous Computing,” \emph{IEEE Trans. Parallel Distrib. Syst.}, vol. 13, no. 3, pp. 260-274, March 2002.
\bibitem{b13} Y. Sahni, J. Cao, L. Yang and Y. Ji, “Multihop offloading of multiple DAG tasks in collaborative edge computing,” \emph{IEEE Internet Things J.}, vol. 8, no. 6, pp. 4893-4905, March, 2021.
\bibitem{b14} Q. Shen, B. -J. Hu and E. Xia, “Dependency-aware task offloading and service caching in vehicular edge computing," \emph{IEEE Trans. Veh. Technol.}, vol. 71, no. 12, pp. 13182-13197, Dec. 2022.
\bibitem{b15} F. Sun et al., “Cooperative Task Scheduling for Computation Offloading in Vehicular Cloud,” \emph{IEEE Trans. Veh. Technol.}, vol. 67, no. 11, pp. 11049-11061, Nov. 2018.
\bibitem{b16} H. Liu, H. Zhao, L. Geng and W. Feng, “A policy gradient based offloading scheme with dependency guarantees for vehicular networks,” in \emph{Proc. IEEE Global Commun. Conf. (GLOBECOM)}, Dec. 2020, pp. 1-6.
\bibitem{b17} Y. Liu, S. Wang, Q. Zhao, S. Du, A. Zhou, X. Ma, and F. Yang, “Dependency-aware task scheduling in vehicular edge computing,” \emph{IEEE Internet Things J.}, pp. 4961–4971, 2020.
\bibitem{b18} J. Shi, J. Du, J. Wang, J. Wang, and J. Yuan, “Priority-aware task offloading in vehicular fog computing based on deep reinforcement learning,” \emph{IEEE Trans. Veh. Technol.}, pp. 16067–16081, 2020.
\bibitem{b19} J. Xie et al., “Advanced Dropout: A model-free methodology for Bayesian dropout optimization," \emph{IEEE Trans. Pattern Anal. Mach. Intell.}, vol. 44, no. 9, pp. 4605-4625, Sept. 2022.
\bibitem{b20} J. Yan, S. Bi and Y. J. A. Zhang, “Offloading and resource allocation with general task graph in mobile edge computing: a deep reinforcement learning approach," \emph{IEEE Trans. Wireless Commun.}, vol. 19, no. 8, pp. 5404-5419, Aug. 2020.
\bibitem{b21} M. S. Mekala et al., “A DRL-based service offloading approach using DAG for edge computational orchestration," \emph{IEEE Trans. Comput. Social Syst.}, Early Access, Apr. 2022.
\bibitem{b22} J. Wang, J. Hu, G. Min, A. Y. Zomaya and N. Georgalas, “Fast adaptive task offloading in edge computing based on meta reinforcement learning," \emph{IEEE Trans. Parallel Distrib. Syst.}, vol. 32, no. 1, pp. 242-253, Jan. 2021. 
\bibitem{b23} M. Goudarzi, M. S. Palaniswami and R. Buyya, “A distributed deep reinforcement learning technique for application placement in edge and fog computing environments," \emph{IEEE Trans. Mobile Comput.}, vol. 22, no. 5, pp. 2491-2505, May 2023.
\bibitem{b24} Z. Hu, J. Tu and B. Li, “Spear: optimized dependency-aware task scheduling with deep reinforcement learning," in \emph{Proc. IEEE Int. Conf. Distrib. Comput. Syst. (ICDCS)}, Oct. 2019, pp. 2037-2046.
\bibitem{b25} X. Wei, L. Cai, N. Wei, P. Zou, J. Zhang and S. Subramaniam, “Joint UAV trajectory planning, DAG task scheduling, and service function deployment based on DRL in UAV-empowered edge computing," \emph{IEEE Internet Things J.}, Early Access, Mar. 2023.
\bibitem{b26} L. Geng, H. Zhao, J. Wang, A. Kaushik, S. Yuan and W. Feng, “Deep reinforcement learning based distributed computation offloading in vehicular edge computing networks," \emph{IEEE Internet Things J.}, Early Access, Feb. 2023.
\bibitem{b27} C. Shu, Z. Zhao, Y. Han, G. Min and H. Duan, “Multi-User Offloading for Edge Computing Networks: A Dependency-Aware and Latency Optimal Approach,” \emph{IEEE Internet Things J.}, vol. 7, no. 3, pp. 1678-1689, March 2020.
\bibitem{b28} M. Taneja and A. Davy, “Resource aware placement of IoT application modules in fog-cloud computing paradigm,” in \emph{Proc. IFIP/IEEE Symp. Integr. Netw. Serv. Manag. (IM)}, Jul. 2017, pp. 1222–1228.
\bibitem{b29} M. Giordani, T. Shimizu, A. Zanella, T. Higuchi, O. Altintas and M. Zorzi, “Path Loss Models for V2V mmWave Communication: Performance Evaluation and Open Challenges," in \emph{Proc. IEEE Connected Automated Vehicles Symp. (CAVS)}, Oct. 2019, pp. 1-5.
\bibitem{b30} Z. Ning, P. Dong, X. Kong, and F. Xia, “A cooperative partial computation offloading scheme for mobile edge computing enabled Internet of thing,” \emph{IEEE Internet Things J.}, vol. 6, no. 3, pp. 4804–4814, Jun. 2019.
\bibitem{b31} P. Velickovic, G. Cucurull, A. Casanova, A. Romero, P. Liò, and Y. Bengio, “Graph Attention Networks," \emph{Proc. Int. Conf. Learn. Representations (ICLR)}, Feb. 2018, pp. 1–12.
\bibitem{b32} Saleh Yousefi, Eitan Altman, Rachid El-Azouzi, and Mahmood Fathy, “Analytical model for connectivity in vehicular ad hoc networks," \emph{IEEE Trans. Veh. Technol.}, vol. 57, no. 6, pp. 3341-3356, Nov. 2008.
\bibitem{b33} S. Misra and S. Bera, “Soft-VAN: Mobility-Aware Task Offloading in Software-Defined Vehicular Network,” \emph{IEEE Trans. Veh. Technol.}, vol. 69, no. 2, pp. 2071-2078, Feb. 2020
\bibitem{b34} Z. He, L. Wang, H. Ye, G. Y. Li and B. -H. F. Juang, “Resource allocation based on graph neural networks in vehicular communications," in \emph{Proc. IEEE Global Commun. Conf. (GLOBECOM)}, Jan. 2020, pp. 1-5.
\bibitem{b35} Y. Li, J. Li, Z. Lv, H. Li, Y. Wang and Z. Xu, “GASTO: A fast adaptive graph learning framework for edge computing empowered task offloading," \emph{IEEE Trans. Netw. Service Manage.}, Early Access, Feb. 2023.
\bibitem{b36} H. Lee, S. Cho, Y. Jang, J. Lee and H. Woo, “A global DAG task scheduler using deep reinforcement learning and graph convolution network," \emph{IEEE Access}, vol. 9, pp. 158548-158561, Nov. 2021.
\bibitem{b37}J. Chen, Y. Yang, C. Wang, H. Zhang, C. Qiu and X. Wang, “Multi-task offloading strategy optimization based on directed acyclic graphs for edge computing," \emph{IEEE Internet Things J.}, vol. 9, no. 12, pp. 9367-9378, Jun. 2022.
\bibitem{b38} M. Liwang, Z. Gao, and X. Wang, “Energy-aware graph task scheduling in software-defined air-ground integrated vehicular networks,” arXiv:2008.01144, 2021. 
\bibitem{b39} J. Wang, C. Jiang, K. Zhang, T. Q. S. Quek, Y. Ren, and L. Hanzo, “Vehicular sensing networks in a smart city: principles, technologies and applications,” \emph{IEEE Wireless Commun.}, vol. 25, no. 1, pp. 122-132, Feb. 2018.
\bibitem{b40} R. J. Williams, “Simple statistical gradient-following algorithms for connectionist reinforcement learning,” Mach. Learn., vol. 8, nos. 3–4, pp. 229–256, 1992.
\bibitem{b41} F. Scarselli, M. Gori, A. C. Tsoi, M. Hagenbuchner, and G. Monfardini, “The Graph Neural Network Model,” \emph{IEEE Trans. Neural Netw.}, vol. 20, no. 1, pp. 61–80, Jan. 2009.
\bibitem{b42} S. Wang, M. Lee, S. Hosseinalipour, R. Morabito, M. Chiang and C. G. Brinton, “Device sampling for heterogeneous federated learning: theory, algorithms, and implementation," in \emph{Proc. IEEE Int. Conf. Comput. Commun. (INFOCOM)}, Jul. 2021, pp. 1-10.
\bibitem{b43} H. Ye, J. Wang and Z. Li, “MIP reformulation for max-min problems in two-stage robust SCUC," \emph{IEEE Trans. Power Syst.}, vol. 32, no. 2, pp. 1237-1247, Mar. 2017.
\bibitem{b44} A. Paszke, S. Gross, F. Massa, A. Lerer, J. Bradbury, G. Chanan, T. Killeen, Z. Lin, N. Gimelshein, L. Antiga, and A. Desmaison, “PyTorch: An imperative style, high-performance deep learning library," in \emph{Proc. Conf. Neural Inf. Process. Syst. (NeurIPS)}, Dec. 2019, pp. 8024–8035.
\bibitem{b45} D. P. Kingma and J. Ba, “Adam: A method for stochastic optimization,” in \emph{Proc. Int. Conf. Learn. Represent. (ICLR)}, Dec. 2015, pp. 1–15.
\bibitem{b46} W. Zhan et al., “Deep-reinforcement-learning-based offloading scheduling for vehicular edge computing," \emph{IEEE Internet Things J.}, vol. 7, no. 6, pp. 5449-5465, Jun. 2020.
\bibitem{b47} X. Yang, H. Luo, Y. Sun and M. Guizani, “A novel hybrid-ARPPO algorithm for dynamic computation offloading in edge computing," \emph{IEEE Internet Things J.}, vol. 9, no. 23, pp. 24065-24078, Dec.1, 2022.
\bibitem{b48} D. Clevert, T. Unterthiner, and S. Hochreiter, “Fast and accurate deep network learning by exponential linear units (ELUs),” in \emph{Proc. Int. Conf. Learn. Represent. (ICLR)}, 2016, pp. 1–14.
\bibitem{b49} V. Mnih, K. Kavukcuoglu, D. Silver, A. A. Rusu, J. Veness, M. G. Bellemare, A. Graves, M. Riedmiller, A. K. Fidjeland, G. Ostrovski, S. Petersen, C. Beattie, A. Sadik, I. Antonoglou, H. King, D. Kumaran, D. Wierstra, S. Legg, and D. Hassabis, “Human-level control through deep reinforcement learning," Nature, vol. 518, no. 7540, pp. 529–533, Feb. 2015.
\bibitem{b50} H. van Hasselt, A. Guez, and D. Silver, “Deep reinforcement learning
with double Q-learning,” in \emph{Proc. AAAI Conf. Artif. Intell.}, Sep. 2016, pp. 2094–2100.
\end{thebibliography}
\end{document}